%% file: main.tex
\documentclass[10pt,twocolumn,letterpaper]{article}

\usepackage[pagenumbers]{cvpr} 


\definecolor{cvprblue}{rgb}{0.21,0.49,0.74}
\usepackage[pagebackref,breaklinks,colorlinks,allcolors=cvprblue]{hyperref}
\usepackage{amsfonts}       
\usepackage{nicefrac}       
\usepackage{multirow}
\usepackage{pifont}
\usepackage{graphicx}
\usepackage{placeins}
\usepackage{floatrow}
\usepackage{caption}
\usepackage{setspace}
\usepackage{amsmath}
\usepackage{amssymb}
\usepackage{amsthm}
\usepackage{bm}
\usepackage{booktabs}
\usepackage{algorithm}
\usepackage{algpseudocode}
\usepackage{xspace}

\newtheorem{proposition}{Proposition}

\newcommand{\method}{SCoPE\xspace}

\newcommand{\ngi}{Normalize-Gate-Inject\xspace}
\newcommand{\dvec}{\mathbf{d}}
\newcommand{\mvec}{\mathbf{m}}
\newcommand{\rvec}{\mathbf{r}}
\newcommand{\ovec}{\mathbf{o}}
\newcommand{\plinner}[2]{\langle #1,\, #2\rangle_{\mathrm{Pl}}}

\providecommand{\Description}[2][]{}

\def\paperID{*****}
\def\confName{CVPR}
\def\confYear{2026}

\newif\ifStoryA
\StoryAfalse   

\ifStoryA
\title{RayPE: Ray-Space Positional Encoding for 3D-Aware Video Generation}
\else
\title{SCoPE: Sightline-Coordinate Positional Encoding for Video Diffusion Transformers}
\fi

\author{
Minghao Yin$^{1*}$ \quad Jiahao Lu$^{2*}$ \quad Wenbo Hu$^{3\dagger}$ \quad Wang Zhao$^3$ \quad Ying Shan$^3$ \quad Kai Han$^{1\ddagger}$
\\[0.3em]
$^1$The University of Hong Kong \quad
$^2$The Hong Kong University of Science and Technology \\
$^3$ARC Lab, Tencent
}

\begin{document}

\twocolumn[{%
\renewcommand\twocolumn[1][]{#1}%
\maketitle

\begin{center}
\vspace{-2em}
\large
Project Page:
\url{https://visual-ai.github.io/scope/}
\end{center}

\ifStoryA
\else
\includegraphics[width=\linewidth]{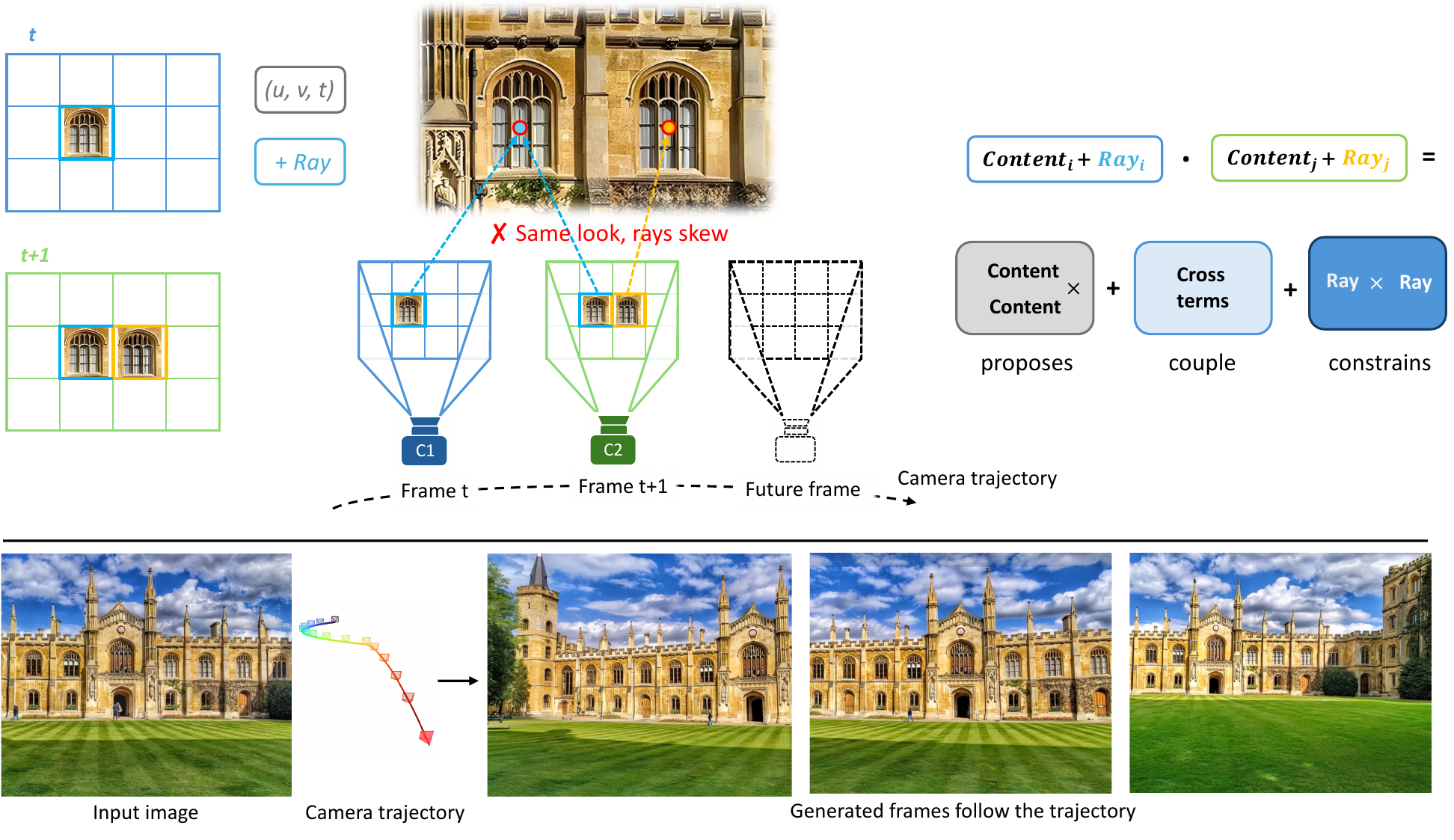}
\captionof{figure}{\textbf{Camera control as a property of the coordinate system.}
    \method gives every token a second address, its camera ray (left).
    Rays are fixed by the user's trajectory while the scene, dashed, is not yet generated (center).
    Appearance cannot choose between the two identical windows in frame $t{+}1$.
    Only one of their rays meets the ray from frame $t$.
    Added to the pretrained queries and keys, the ray gives the score exactly this test, zero when rays meet, growing with the miss (right).
    %
    Content proposes matches, ray pairing constrains compatible sightlines, and cross terms couple token features with camera rays.
    }
\fi
\label{fig:teaser}
\vspace{1em}
}]

\begingroup
\renewcommand{\thefootnote}{}
\footnotetext{* Equal contribution \quad $^\dagger$ Project lead. \quad $^\ddagger$ Corresponding author.}
\endgroup

\ifStoryA
\else
\input{sections/00_abstract}
\fi


\ifStoryA
\else
\input{sections/01_introduction}
\fi
\input{sections/02_related_work}
\input{sections/03_background}
\input{sections/04_method}
\input{sections/06_experiments}

\input{sections/08_conclusion}

\clearpage

{
    \small
    \bibliographystyle{ieeenat_fullname}
    \bibliography{references}
}

\clearpage
\appendix
\input{sections/A_appendix}
\input{sections/05_training}

\end{document}

%% file: sections/00_abstract_storyA.tex
\begin{abstract}

Modern video diffusion transformers position their tokens through RoPE on the $(u,v,t)$ axes --- a description of the camera's sampling grid that says nothing about the 3D structure of the scene. 
We observe that the geometric relation between two camera rays is captured by the Pl\"ucker reciprocal product, which is bilinear in the two rays --- the same algebraic form as the dot product in Transformer attention. 
Building on this analogy, we propose \method, a positional-encoding extension that injects per-token 6D Pl\"ucker coordinates additively into the queries and keys of self-attention, with a query/key flip arrangement under which the symmetric identity
configuration coincides exactly with the reciprocal product. 
The injection is additive, the resulting attention score decomposes into a content term, a geometry term, and two content$\leftrightarrow$ geometry cross-terms --- all of which our experiments find individually necessary. 
To make the encoding stable across video data with heterogeneous camera-translation scales (SfM, deep SLAM, metric), we further decouple ray direction from moment magnitude, gate the encoding by a learned function of the log-magnitude, and
apply RMSNorm to align it with the QKNorm-normalized content branch.
aThe full module adds less than $0.1\%$ parameters to a pretrained video DiT, is zero-initialized to start from the pretrained weights, and improves camera controllability, cross-frame 3D consistency, and overall video quality on a four-dataset training mixture.

\end{abstract}

%% file: sections/00_abstract.tex
\begin{abstract}
Video diffusion transformers address their tokens by position on the pixel-time grid: an address in the tensor, not in the world.
The address we would want, the world point a token depicts, lies on a surface not yet generated, while its camera ray is fixed once the user specifies a trajectory.
\method therefore treats the ray as a second positional coordinate, and camera control becomes a property of the coordinate system, not an added module.
The ray is added to the pretrained attention's queries and keys, and the score gains a term that reads the two rays alone.
Its canonical form, the reciprocal product of line geometry, measures how nearly two lines of sight meet.
Normalize-Gate-Inject makes a single encoding trainable across metric and up-to-scale pose sources.
The retrofit keeps RoPE bit-exact, starts from the unchanged pretrained DiT, and adds under $0.1\%$ new parameters.
%
On Wan2.2 at 5B and 14B under matched data and budget, \method improves every camera-controllability and fidelity metric, leads all closed-loop revisit metrics, and shows widening margins with model size.
At 14B, rotation error falls $29\%$ and FVD $43\%$ below the strongest baseline.
\end{abstract}

%% file: sections/01_introduction_storyA.tex
\section{Introduction}
\label{sec:intro}

Video diffusion transformers have become the dominant backbone for
large-scale text-to-video and image-to-video
generation~\cite{ho2022vdm,blattmann2023svd,yang2025cogvideox,polyak2024moviegen,kong2024hunyuanvideo, wan2025wan}.
These models patchify each frame into a grid of latent tokens and
inject positional information through a factorized RoPE on the
$(u,v,t)$ index~\cite{su2021rope,heo2024rotary}. This positional
encoding describes the \emph{sampling grid} of the camera: it tells
the model where each token sits in the 2D lattice and along the
temporal axis, but says nothing about \emph{where in 3D} the
corresponding ray points. Two tokens that observe the same physical
surface from different viewpoints carry no positional relationship, and
the model must recover their geometric correspondence entirely from
pixel content.

Recent camera-conditioned video
models~\cite{he2025cameractrl,wang2024motionctrl,xu2024camco,yu2024viewcrafter,yu2025trajectorycrafter,bai2025recammaster}
address this gap by injecting camera pose through adapter modules,
ControlNet encoders, rendered point-cloud images, or cross-attention
streams. These approaches treat geometry as an auxiliary feature
processed outside of self-attention, while the attention dot product
remains purely content-based. A parallel line of
work~\cite{li2025prope,zhang2026ucpe,li2026rerope} attempts to build
camera information into positional encodings, but does so
multiplicatively by replacing or splitting the original RoPE, which
disrupts the pretrained positional structure and employs reduced
parameterizations that do not exploit the algebraic structure of
attention.

We approach the problem from a different angle. A video pixel can be
identified with the camera ray that produced it, parameterized as a 6D
Pl\"ucker coordinate. The geometric relationship between two rays is
captured by the Pl\"ucker reciprocal product (the Klein form on line
space), which is bilinear in the two rays, SE(3)-invariant, and
vanishes exactly when the rays are coplanar (i.e., observe the same 3D
point). Transformer attention computes a dot product between queries
and keys, which is also a bilinear form. These two share the same
algebraic shape. We ask: can the geometric relationship between two
camera rays be made to live \emph{inside} the attention dot product, so
that the model receives 3D information through the same channel that
already carries positional and content signals?

\begin{figure}[!t]
  \centering
  \includegraphics[width=\linewidth]{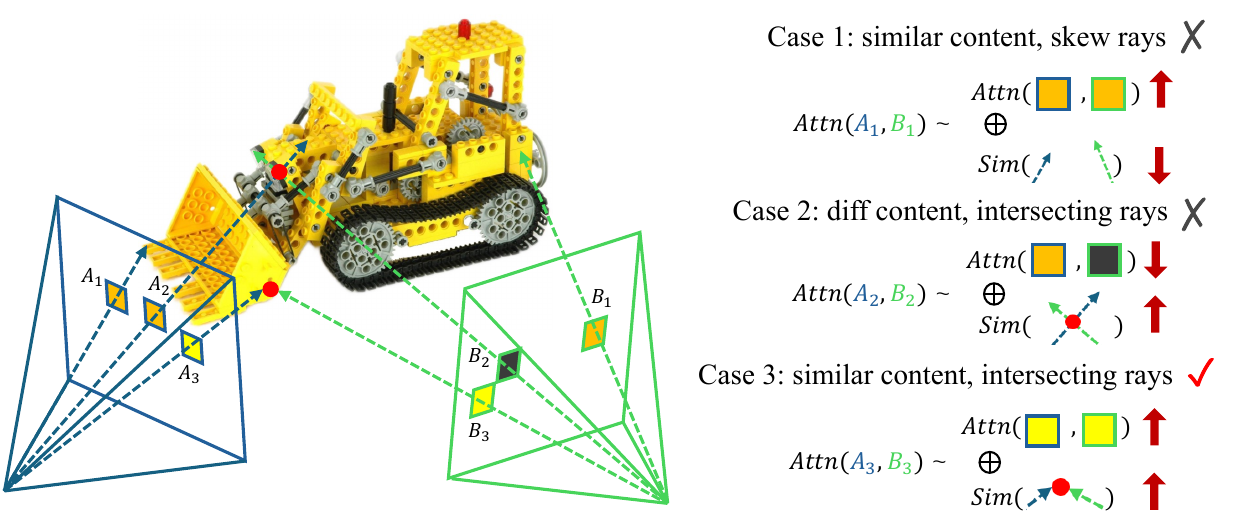}
  \caption{Why attention needs ray geometry. Two cameras (blue,
    green) each cast three rays through the tokens
    $A_{1,2,3}$ and $B_{1,2,3}$ on their image planes; red dots mark
    where two rays meet in 3D. Whether a pair $(A_*,B_*)$ should
    attend factors as content similarity and ray-geometric
    similarity. \textbf{Case~1} (similar content, skew rays) and
    \textbf{Case~2} (different content, intersecting rays) are both
    misleading; only \textbf{Case~3} (both agree) is a genuine
    correspondence. \method realizes this product inside the
    attention dot product by injecting Pl\"ucker rays into $Q,K$
    (Sec.~\ref{sec:method-flip}).}
  \Description{Two pinhole cameras (blue on the left, green on the
    right) view a yellow toy bulldozer; each image plane carries
    three labeled token patches, with dashed rays from the camera
    centre through each patch, and red dots marking where pairs of
    rays meet in 3D. The right column lists three cases that factor
    the attention between a token pair into a content-similarity
    factor and a ray-geometric-similarity factor, with each case
    annotated by a check or a cross.}
  \label{fig:concept}
\end{figure}

We answer affirmatively with \method (\emph{Ray-Space Positional
Encoding}). For each token we compute its 6D Pl\"ucker coordinate
from the camera pose and intrinsics, project it through a small
learned head, and add the result to the query and key vectors of
self-attention. The injection is additive and zero-initialized,
leaving the pretrained 3D RoPE intact at the start of training. The
resulting attention score decomposes into a content term, a geometry
term, and two content$\leftrightarrow$geometry cross-terms; our
ablations show all three components are individually necessary.

Two practical issues must be solved before this encoding works at
scale. First, the Pl\"ucker moment scales linearly with camera
translation, so datasets with different pose conventions (SfM~\cite{schoenberger2016sfm},
SLAM~\cite{durrant2006simultaneous,teed2021droidslam}, metric) produce wildly different encoding magnitudes. Second,
naively adding the geometric signal to the content branch does not
control their relative magnitude. We address both through
\emph{Normalize-Gate-Inject} (NGI): we decouple the ray into a
scale-invariant direction and a log-magnitude, gate the encoding
by a learned function of the log-magnitude, and apply an RMSNorm~\cite{zhang2019root}
symmetric to the content-branch QKNorm~\cite{henry2020qknorm}. The result is stable across
heterogeneous data while preserving absolute distance information.

We integrate \method into Wan2.2-TI2V-5B~\cite{wan2025wan}, a 5B-parameter pretrained
video diffusion transformer, and train on a mixture of four datasets
(RealEstate10K~\cite{zhou2018re10k}, DL3DV~\cite{ling2024dl3dv}, PanShot~\cite{zhang2026ucpe}, and
OmniWorld~\cite{zhou2025omniworld}). The encoding is zero-initialized so training starts exactly
from the pretrained model, the original RoPE is preserved, and the
total parameter increase is below $0.1\%$. Experimental results on
camera controllability, cross-frame 3D consistency, and video
generation quality are reported in Section~\ref{sec:experiments}.
Our contributions center on two ideas and their practical realization:
\begin{itemize}

  \item \textbf{Pl\"ucker rays as positional encoding.} The Pl\"ucker reciprocal product is bilinear in two rays --- the same algebraic form as the Transformer attention score --- so camera rays admit a PE-style injection inside the attention dot product, not around it.
  \item \textbf{\method.} An additive 6D Pl\"ucker injection on $q$ and $k$ that splits attention into a content term, a geometry term, and two content$\leftrightarrow$geometry cross-terms. Ablations show all three are individually necessary; the cross-terms are a coupling that V-side, cross-attention, AdaLN, and rotation-based ray PEs cannot structurally represent.
  \item \textbf{Normalize-Gate-Inject.} A decoupling, gating, and PE-side RMSNorm pipeline that stabilizes 6D Pl\"ucker injection across video data with heterogeneous pose-scale conventions (SfM, SLAM, metric).
\end{itemize}

%% file: sections/01_introduction.tex
\section{Introduction}
\label{sec:intro}

A transformer knows only as much about space as its positional encoding supplies.
In video diffusion transformers (DiTs)~\cite{ho2022vdm,yang2025cogvideox,kong2024hunyuanvideo,wan2025wan}, that encoding is typically a factorized RoPE~\cite{su2021rope,heo2024rotary} over the $(u,v,t)$ pixel-time grid: \emph{an address in the tensor, not in the world}.
Two tokens imaging the same surface from different viewpoints therefore receive positions related only by grid offset, and the model must recover 3D correspondence from appearance alone.

Ideally, each token would sit at the world point it depicts, but that point is where its line of sight meets a surface the model has yet to generate.
World-point methods must first reconstruct existing content~\cite{bai2025pefield,xu2026ucm,wang2026warphistory,yu2024viewcrafter}, and new content, as in text-to-video, offers nothing to reconstruct.
The line of sight, the token's ray, is another matter.
\emph{A ray belongs to the camera, not to the content}.
Once the user specifies a trajectory~\cite{he2025cameractrl,bai2025recammaster,burgert2025flow}, every future token's ray is fixed before a single pixel is synthesized.
A ray is admittedly less than a point, fixing the line of sight but not the depth along it.
Yet it supplies the constraint appearance alone cannot, since two tokens can depict the same point only if their rays nearly meet (Figure~\ref{fig:teaser}).
We therefore give each token its ray as a second address, and camera control stops being an auxiliary signal bolted onto the model and becomes \emph{a property of the coordinate system itself.}

A coordinate, though, is just half of a positional encoding.
\emph{The other half is the rule by which a pair of coordinates enters the attention score}.
On the pixel-time grid, that rule is RoPE~\cite{su2021rope}, whose rotations collapse two absolute indices into their offset, the grid's relative quantity.
For camera rays, a line of work running from multi-view synthesis~\cite{miyato2024gta,kong2024eschernet,li2025prope} into video diffusion~\cite{zhang2026ucpe,li2026rerope,xiang2026viewrope} extends this recipe, applying relative camera transforms \emph{multiplicatively} to queries and keys.
We build on this line and part ways with it over one question.
\emph{What must the geometric score measure?}
Deciding whether two tokens could see the same point makes three demands of the score: (i) it must relate individual tokens, not frames; (ii) it must register camera translation, since parallax lives in the baseline; and (iii) it must stay readable from the two rays alone, so one test serves any content.
Each demand rules out an established design.
Frame-level transforms give every token in a frame the same geometry (i)~\cite{miyato2024gta,kong2024eschernet,li2025prope,li2026rerope}.
Direction-only encodings are blind to translation (ii)~\cite{xiang2026viewrope}.
Multiplicative application routes geometry entirely through content, leaving the score with no term that reads the two rays by themselves (iii)~\cite{zhang2026ucpe}.
The first two fix the coordinate, and the third asks for a different way into the score.

We present \method (Sightline-Coordinate Positional Encoding), which meets all three demands with \emph{a single addition}.
Each token's full Pl\"ucker ray \(\rvec\), whose moment carries the translation, is added to its query and key after the pretrained RoPE.
Because the ray enters as a summand rather than a transform, the dot product expands into four terms: content with content, the two content--ray cross terms, and ray with ray.
The last term reads the two rays and no content at all, the pairing that demand (iii) calls for.
Unlike grid indices, rays admit no subtraction, so their relative quantity is not an offset but incidence.
Line geometry's classical incidence measure is the \emph{reciprocal product}, bilinear in the two rays and exactly the shape of the new term.
It vanishes when two rays meet, or more generally lie in one plane, and grows with how far they miss.
%
%
Learned projections lift each ray into query-key space, so the pairing is a bilinear form, not a hard-wired formula.
It reduces to the reciprocal product when the projections match each ray's direction with the other's moment.
The content term identifies candidate matches from visual evidence,
while the ray pairing filters matches with compatible sightlines
(Figure~\ref{fig:teaser}).
With the ray as a second coordinate, a token becomes a joint state
\(z=(x,\rvec)\), combining its learned feature \(x\) with its
prescribed camera ray \(\rvec\).
Scoring content--content and ray--ray compatibility independently
imposes a separable score on this joint token state.
The cross terms remove this restriction by introducing direct
interactions between visual content and viewing geometry.


The moment that carries the translation, however, also carries its scale, set by whatever produced the poses.
Poses from metric capture come in meters, while monocular SfM~\cite{schoenberger2016sfm} and SLAM~\cite{durrant2006simultaneous,teed2021droidslam} recover translation only up to scale, so their unit is a per-clip convention, not a measurement.
Prior camera encodings normalize offline to one fixed rule, which costs nothing where the unit is arbitrary.
In camera-controlled generation, though, scale is part of the command, and a centimeter dolly should not look like a ten-meter sweep.
\method keeps that scale by splitting each ray into a scale-invariant part and an explicit log magnitude, so a change of unit moves one number, not the whole feature.
A learned gate reads that magnitude and sets the injection strength for each token, so no single weight serves every convention.
Because the backbone normalizes its queries and keys, the injected features receive a matching RMSNorm~\cite{zhang2019root} and enter the score on equal footing.
Together, these steps normalize the ray, gate its influence, and inject it, hence the name \ngi.
The additions total under $0.1\%$ of the parameters, leave RoPE bit-exact, and add no second attention or adapter branch.
The injection is zero at initialization, so training starts from the exact pretrained DiT.

If camera control really is a property of the coordinate system, three things should follow.
Adding the ray coordinate, and changing nothing else, should improve control.
The improvement should trace to the pairing term the derivation singled out.
And it should hold as the backbone grows.
We test all three by retrofitting \method onto Wan2.2~\cite{wan2025wan} at 5B and 14B.
Every compared method shares the backbone, data, and training budget, so the camera conditioning is the only thing that differs.
All three predictions hold.
Control and fidelity improve together.
\method leads on every camera-controllability metric and on FVD, FVD$_{\mathrm{c}}$, and FID.
On closed-loop camera trajectories, it also leads all revisit metrics, recovering both the initial camera pose and the conditioning view after the commanded excursion.
At 14B, rotation error falls $29\%$ and FVD $43\%$ below the strongest baseline on each metric.
Per-term ablations put the largest cost on removing the ray pairing, the very term the multiplicative form does not produce.
Without it, rotation error nearly doubles.
From 5B to 14B, the margins widen rather than wash out, and the translation-error lead grows from $12\%$ to $25\%$.

\begin{itemize}
  \item \textbf{Rays as positional coordinates.}
    A single zero-initialized addition puts each token's Pl\"ucker ray into the pretrained queries and keys, giving the score a pure ray pairing whose canonical form is the reciprocal product.
    %
  \item \textbf{Normalize-Gate-Inject.}
    A scale-invariant split, a per-clip learned gate, and a matched norm confine arbitrary pose units to one number and keep the commanded scale in the encoding.
  \item \textbf{Controlled evidence at scale.}
    Under matched backbone, data, and budget, \method improves every camera-controllability and fidelity metric, the gains trace to the ray pairing, and the margins grow with model size.
\end{itemize}

%% file: sections/02_related_work.tex
\section{Related Work}
\label{sec:related}

\noindent\textbf{Camera control by conditioning.}
Adapter methods attach modules outside the pretrained attention, encoding extrinsics as per-frame tokens~\cite{wang2024motionctrl}, rasterizing per-pixel Pl\"ucker maps into a ControlNet-style branch~\cite{he2025cameractrl}, concatenating camera embeddings with epipolar cross-view attention~\cite{xu2024camco}, or conditioning trainable cross-attention and LoRA modules~\cite{bahmani2025ac3d,bahmani2025vd3d}.
Successors condition directly on pose and time~\cite{vanhoorick2024gcd,watson20254dim}, steer trajectory or dimension adapters~\cite{yang2024directavideo,xiao2025trajectoryattention,sun2025dimensionx}, and scale the recipe to dynamic scene exploration~\cite{he2025cameractrlii,huang2026spacetimepilot}.
Rendering methods reconstruct a proxy and re-render it along the target trajectory as a pixel-aligned condition~\cite{yu2024viewcrafter,yu2025trajectorycrafter,gu2025diffusion,bian2025gsdit,ren2025gen3c,li2025realcami2v,lee2026pointtracks}, or steer a frozen model training-free through warped views and rearranged point clouds~\cite{you2025nvssolver,hou2025camtrol,song2026worldforge}.
ReCamMaster and multi-camera variants concatenate view tokens and let the native attention discover correspondence on its own~\cite{bai2025recammaster,bai2025syncammaster,kuang2024cvd,xu2025cavia}, while Go-with-the-Flow and Latent-Reframe move the camera into noise and latents, a route orthogonal to ours and combinable with it~\cite{burgert2025flow,zhou2025latentreframe}.
These routes pay for control in added modules, extra sequence length, or an upstream reconstruction stage.
A related line addresses tokens by world-space coordinates estimated from existing content~\cite{bai2025pefield,xu2026ucm,wang2026warphistory}.
The ray needs no reconstruction and is fixed by the trajectory before generation begins.

\smallskip\noindent\textbf{Camera geometry inside attention.}
A more direct line moves the camera into the attention operator itself.
Epipolar attention did so first as a mask, constraining which pairs may attend~\cite{he2020epipolar,du2023widebaseline}.
GTA~\cite{miyato2024gta} made the geometry an operator instead, applying relative SE(3) transforms multiplicatively to the queries, keys, and values of a multi-view transformer trained from scratch.
EscherNet's camera encoding and RePAST follow the same pattern~\cite{kong2024eschernet,safin2023repast}.
PRoPE generalizes the transform from extrinsics to full camera frustums while keeping the patch RoPE on separate channels~\cite{li2025prope}, and interactive world models have already adopted it~\cite{sun2025worldplay}.
In video diffusion, UCPE runs GTA-style operators in a parallel zero-initialized bypass branch, at roughly $0.5\%$ added parameters and one extra attention per block~\cite{zhang2026ucpe}.
ReRoPE overwrites the low-frequency temporal RoPE phases of a frozen backbone with lifted relative rotations~\cite{li2026rerope}.
ViewRope rotates query and key subvectors by viewing-ray direction on the same backbone we use~\cite{xiang2026viewrope}.
In each of these, the geometry enters multiplicatively and is routed through content, so the score gains no term that depends on the two rays alone.
Scale gets at most a fixed offline rule, near-plane normalization in UCPE, single-convention data in GTA and PRoPE, nothing in ReRoPE.
\method stays in this line and changes the geometry's role, from a transform on content to a coordinate of the token, added in the pretrained attention with no branch, no second attention, and a learned gate instead of a fixed rule.
The line keeps moving.
ReDirector folds cameras into RoPE phase shifts~\cite{park2026redirector}, URoPE carries relative embeddings across geometric spaces~\cite{xie2026urope}, and DPPE decouples rotation from translation for scaling~\cite{kenney2026dppe}.
CRePE adds depth distributions to UCPE's rays, an upgrade the additive form can adopt unchanged~\cite{jin2026crepe}, CameraAnything applies ReDirector's rotary phases to intrinsic and extrinsic editing~\cite{li2026cameraanything}, and RayRoPE, no relation despite the name, targets feedforward reconstruction~\cite{wu2026rayrope}.

\smallskip\noindent\textbf{Positional encodings and ray-space representations.}
Relative positional encodings enter the score as additive terms in Transformer-XL and T5~\cite{dai2019transformerxl,raffel2020t5}, or as rotations that collapse indices into offsets in RoPE and its factorized video form~\cite{su2021rope,heo2024rotary}.
\method extends the additive family to a coordinate whose relative quantity is incidence rather than offset.
Rays are also an established representational domain.
The plenoptic function and light-field rendering parameterize appearance over rays~\cite{adelson1991plenoptic,levoy1996light,gortler1996lumigraph}, neural fields from NeRF to light field networks and light-field transformers learn scene functions on ray inputs~\cite{mildenhall2020nerf,sitzmann2021lfns,wang2022r2l,suhail2022light,attal2022learning}, and SE(3) equivariant transformers operate in ray space~\cite{xu2023rayspace}.
Camera pose can be estimated as a bundle of rays~\cite{zhang2024camerasasrays}, world models generate autoregressively in ray space~\cite{xie2026raynova}, and Rays as Pixels folds camera trajectories into the video distribution~\cite{jang2026raysaspixels}.
These works treat the ray as what the network represents or predicts.
\method treats it as where a token stands, a coordinate the pretrained attention compares.
Retrofitting favors additions that vanish at initialization, the principle behind ControlNet's zero-initialized residuals~\cite{zhang2023controlnet}, and \method carries it into the positional encoding.

%% file: sections/03_background.tex
\section{Background and Notation}
\label{sec:background}

This section fixes the notation we use for video diffusion
transformers, the camera model, and the Pl\"ucker representation of a
ray, and recalls the geometric facts that motivate our design.

\subsection{Video Diffusion Transformers}
\label{sec:bg-dit}

We follow the latent-video diffusion / flow-matching
setup~\cite{ho2020ddpm,lipman2023flow,blattmann2023svd,yang2025cogvideox,wan2025wan}.
A video $V\in\mathbb{R}^{T\times 3\times H\times W}$ is encoded by a
3D VAE into a latent volume $Z\in\mathbb{R}^{T_z\times C\times H_z\times W_z}$,
patchified into a sequence of tokens
$X\in\mathbb{R}^{N\times d}$ with $N=T_z\!\cdot\!H_p\!\cdot\!W_p$,
where $H_p,W_p$ are the post-patchification spatial sizes. Each token
carries an integer index $(u,v,t)\in\{0,\dots,W_p-1\}\times\{0,\dots,H_p-1\}
\times\{0,\dots,T_z-1\}$.

A self-attention layer~\cite{vaswani2017attention} with query/key/value
projections $W_Q,W_K,W_V\!\in\!\mathbb{R}^{d\times d}$ produces, for
each token $i$,
\begin{equation}
  q_i = W_Q\,x_i,\quad k_i = W_K\,x_i,\quad v_i = W_V\,x_i,
\end{equation}
and computes the attention output
$o_i = \sum_j \mathrm{softmax}_j\!\bigl(\tfrac{1}{\sqrt{d}}\langle q_i,k_j\rangle\bigr)\,v_j$.
Modern video DiTs add (i) RMS-based query/key
normalization~\cite{henry2020qknorm} (often called QKNorm) and (ii) a
factorized 3D rotary positional encoding (RoPE) that applies a per-pair
rotation to $q$ and $k$ as a function of $(u,v,t)$:
\begin{equation}
  \tilde q_i = R(u_i,v_i,t_i)\,q_i,\quad
  \tilde k_j = R(u_j,v_j,t_j)\,k_j,
  \label{eq:rope}
\end{equation}
where $R(\cdot)$ is a block-diagonal rotation built from per-axis 1D
RoPE~\cite{su2021rope,heo2024rotary}. The rotation is applied so that
the dot product $\langle \tilde q_i,\tilde k_j\rangle$ depends on
$(u_i\!-\!u_j, v_i\!-\!v_j, t_i\!-\!t_j)$, giving the standard
relative positional bias on the sampling lattice.

Crucially, $(u,v,t)$ are indices into the camera's sampling grid; they
say nothing about \emph{where} in 3D each token's pixel originated. We
do not change Eq.~\eqref{eq:rope}; \method adds an additive term
that is computed from the 3D camera ray of each token.

\subsection{Camera Model and Per-Token Rays}
\label{sec:bg-camera}

Each video clip is annotated with per-frame extrinsics
$T_t\!\in\!\mathrm{SE}(3)$ (camera-to-world) and per-frame intrinsics
$K_t\!\in\!\mathbb{R}^{3\times 3}$. We adopt the OpenCV convention
($x$-right, $y$-down, $z$-forward) throughout. For a token whose
patch center has pixel coordinate $(p_u,p_v)$ in frame $t$, the
camera-frame ray direction is
\begin{equation}
  \dvec^{\mathrm{cam}}_i \;=\;
  \frac{K_t^{-1}\,[p_u,p_v,1]^\top}
       {\|K_t^{-1}\,[p_u,p_v,1]^\top\|},
\end{equation}
the world-frame ray direction is $\dvec_i = R_t\,\dvec^{\mathrm{cam}}_i$
where $R_t$ is the rotation part of $T_t$, and the world-frame ray
origin is $\ovec_i$, the position of camera $t$. By construction
$\|\dvec_i\|=1$.

\subsection{Pl\"ucker Decomposition Coordinates and Reciprocal Product}
\label{sec:bg-plucker}

A 3D line (or ray) in $\mathbb{R}^3$ is identified by a pair of
3-vectors $(\dvec,\mvec)\in\mathbb{R}^6$, where $\dvec$ is the
direction and
\begin{equation}
  \mvec \;=\; \ovec \times \dvec
  \label{eq:plucker-moment}
\end{equation}
is the Pl\"ucker moment, with $\ovec$ any point on the line. Two
constraints make $(\dvec,\mvec)$ a parameterization of an oriented
line up to scale:
$\|\dvec\|>0$ and $\dvec\cdot\mvec=0$. Both are direct consequences of
Eq.~\eqref{eq:plucker-moment}~\cite{hartley2004multi,pottmann2001computational}.

Given two rays $\rvec_i=(\dvec_i,\mvec_i)$ and
$\rvec_j=(\dvec_j,\mvec_j)$, their \emph{Pl\"ucker reciprocal product}
(also called Pl\"ucker inner product) is

\begin{equation}
\begin{aligned}
\plinner{\rvec_i}{\rvec_j}
&\equiv \dvec_i \,\cdot\, \mvec_j + \dvec_j \,\cdot\, \mvec_i \\
&= \rvec_i^\top J\,\rvec_j,
\quad
J = \begin{pmatrix}\mathbf{0}_3 & I_3 \\ I_3 & \mathbf{0}_3\end{pmatrix}.
\end{aligned}
\label{eq:plucker-inner}
\end{equation}

Three properties of Eq.~\eqref{eq:plucker-inner} are central to our
design~\cite{hartley2004multi,pottmann2001computational}:

\begin{itemize}
  \item \textbf{Bilinearity.}
    $\plinner{\rvec_i}{\rvec_j}$ is linear in $\rvec_i$ and in
    $\rvec_j$ separately; this is the same algebraic shape as the
    attention dot product $\langle q_i,k_j\rangle$, which is linear
    in $q$ and $k$ separately.
  \item \textbf{Coplanarity criterion.}
    $\plinner{\rvec_i}{\rvec_j}=0$ if and only if the two rays are
    coplanar; equivalently they intersect, are parallel, or coincide.
    For two rays that observe the \emph{same} 3D point through two
    different cameras, the two rays meet at that point and so are
    coplanar, hence $\plinner{\rvec_i}{\rvec_j}=0$.
  \item \textbf{SE(3) invariance.} For $g\in\mathrm{SE}(3)$ acting
    jointly on $\rvec_i$ and $\rvec_j$,
    $\plinner{g\!\cdot\!\rvec_i}{g\!\cdot\!\rvec_j}=\plinner{\rvec_i}{\rvec_j}$;
    the reciprocal product depends only on the relative geometry of
    the two rays, not on the choice of world frame.
\end{itemize}

The bilinear shape of Eq.~\eqref{eq:plucker-inner} and the
bilinear shape of $\langle q,k\rangle$ are the algebraic match that
\method exploits.

\subsection{Scale Behavior of Pl\"ucker Coordinates}
\label{sec:bg-scale}

A subtlety of Pl\"ucker coordinates that becomes important in
practice is their behavior under rescaling of the scene. Suppose all
camera positions are rescaled by a factor $s>0$:
$\ovec\!\to\!s\ovec$. Then $\dvec$ is unchanged (it is a unit
direction) while $\mvec=\ovec\times\dvec\!\to\!s\mvec$. Substituting
into Eq.~\eqref{eq:plucker-inner} gives
\begin{equation}
  \plinner{\rvec_i}{\rvec_j}\!\to\! s\cdot\plinner{\rvec_i}{\rvec_j},
\end{equation}
so the reciprocal product scales linearly with translation magnitude.
Different videos in the wild come with very different translation
scales --- normalized scene volumes, metric meters, or DROID-SLAM's
internal scale~\cite{teed2021droidslam} --- and a naive raw-Pl\"ucker
encoding would produce wildly different geometry-channel magnitudes
across the dataset. Section~\ref{sec:method-ngi} introduces our
\ngi step that is designed to address exactly this issue.

%% file: sections/04_method.tex
\section{Method}
\label{sec:method}
\begin{figure}[t]
  \centering
  \includegraphics[width=\textwidth]{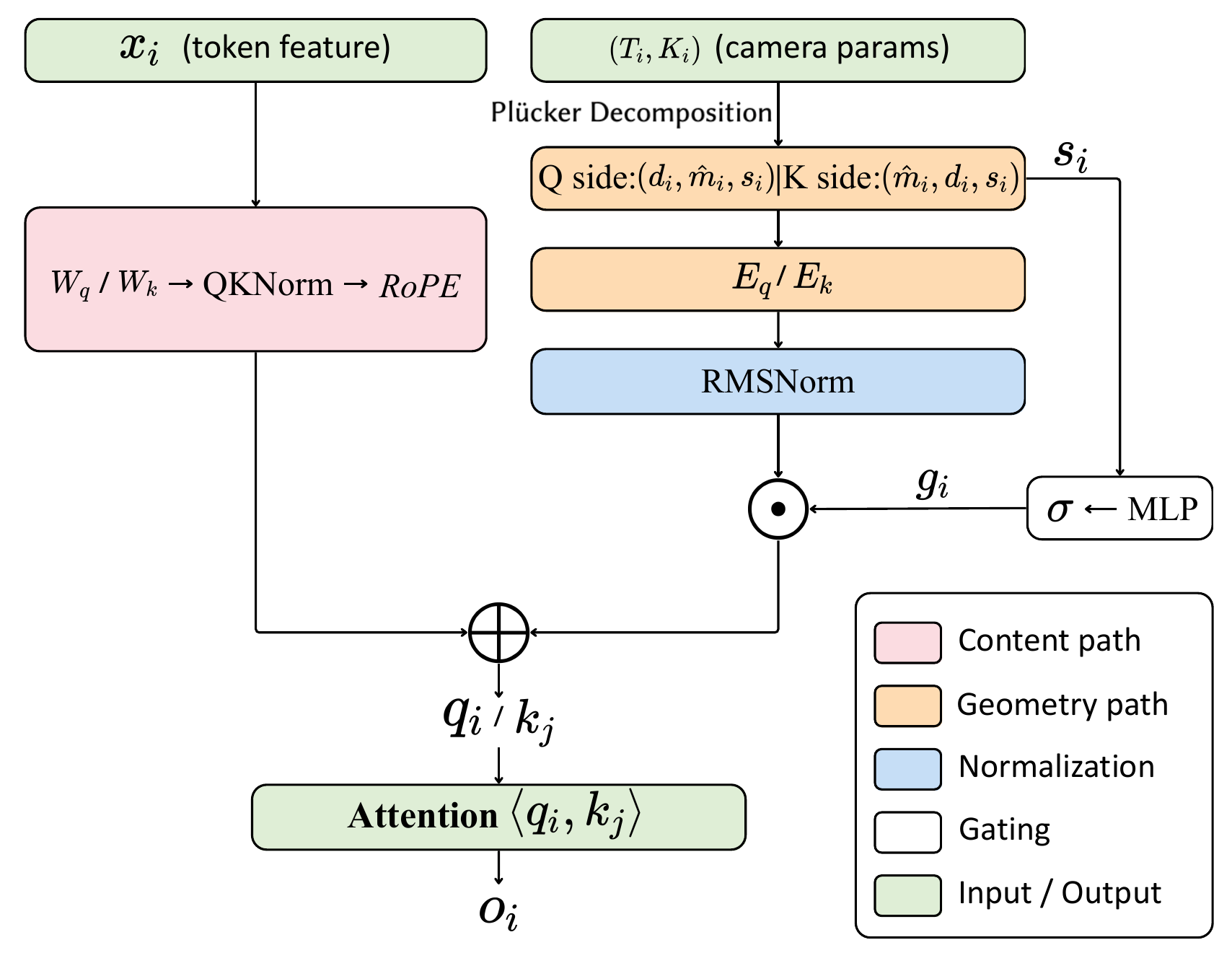}
  \caption{
\textbf{Applying \method to a self-attention layer.}
The content path (pink) preserves the pretrained
$W_q/W_k \rightarrow \mathrm{QKNorm} \rightarrow \mathrm{RoPE}$
pipeline unchanged.
The geometry path (orange/blue) computes per-token Pl\"ucker coordinates
from camera parameters, decomposes them into scale-invariant direction and
log-magnitude, projects the resulting features through \(E_q/E_k\), normalizes
($\mathrm{RMSNorm}$), and gates by a learned function of the log-magnitude.
}
  \Description{}
  \label{fig:attention}
\end{figure}


\method adds a ray-derived positional term to the self-attention
layers of a pretrained video diffusion transformer, as illustrated
in Figure~\ref{fig:attention}. The method has two parts. First,
Pl\"ucker Flip PE injects each token's camera ray additively into the
query and key, preserving the pretrained RoPE while introducing
ray--ray and mixed feature--ray terms into the attention score
(Section~\ref{sec:method-flip}). Second, Normalize--Gate--Inject
controls the scale and magnitude of this added signal across
heterogeneous pose sources (Section~\ref{sec:method-ngi}).
Implementation details are given in
Section~\ref{sec:method-impl}.
\subsection{Pl\"ucker Flip Positional Encoding}
\label{sec:method-flip}

The key observation is that the Pl\"ucker reciprocal product and the
attention score have the same algebraic form. For two rays
$\rvec_i=(\dvec_i,\mvec_i)$ and $\rvec_j=(\dvec_j,\mvec_j)$,
\begin{equation}
\langle \mathbf{r}_i, \mathbf{r}_j \rangle_{\mathrm{Pl}}
=
\mathbf{d}_i \cdot \mathbf{m}_j
+
\mathbf{d}_j \cdot \mathbf{m}_i
\end{equation}
is bilinear in the two rays, while the attention score
$\langle q_i,k_j\rangle$ is bilinear in query and key. This suggests
adding ray-derived features to $q$ and $k$ in an order that makes the
geometric reciprocal product appear as part of the attention score.

Each token $i$ has a camera ray with Pl\"ucker coordinate
$\rvec_i=(\dvec_i,\mvec_i)\in\mathbb{R}^6$
(Section~\ref{sec:bg-camera}). We use two per-layer projections
$E_q,E_k\in\mathbb{R}^{d\times 6}$ and add the projected ray features
after the usual content projection and RoPE:
\begin{align}
  q_i &= R(u_i,v_i,t_i)\,W_Q x_i \;+\; E_q\,(\dvec_i,\,\mvec_i),
  \label{eq:flip-q}\\
  k_j &= R(u_j,v_j,t_j)\,W_K x_j \;+\; E_k\,(\mvec_j,\,\dvec_j).
  \label{eq:flip-k}
\end{align}
Here we use the shorthand
\(\tilde W_Q x_i \equiv R(u_i,v_i,t_i)W_Qx_i\) and
\(\tilde W_K x_j \equiv R(u_j,v_j,t_j)W_Kx_j\) for the
RoPE-encoded content query and key, respectively.
Two choices matter here. The geometry term is additive, so the
pretrained $(u,v,t)$ RoPE remains unchanged. The Pl\"ucker halves are
also flipped between the two sides: the query receives
$(\dvec,\mvec)$, while the key receives $(\mvec,\dvec)$. We discuss
the role of this flip below; this subsection uses the raw 6D
coordinate to make the algebra explicit, and
Section~\ref{sec:method-ngi} gives the normalized 7D feature used in
practice.

\paragraph{Attention score decomposition.}
\label{sec:method-decomp}
Expanding $\langle q_i, k_j\rangle$ using
Eqs.~\eqref{eq:flip-q}--\eqref{eq:flip-k} yields four terms:
\begin{equation}
\begin{aligned}
\langle q_i,k_j\rangle
=&\;
\underbrace{\langle \tilde W_Q x_i,\,\tilde W_K x_j\rangle}_{\text{(A) content}}
+
\underbrace{\langle \tilde W_Q x_i,\,E_k\tilde\rvec_j\rangle}_{\text{(B) content$\to$geom}} \\
&+
\underbrace{\langle E_q\rvec_i,\,\tilde W_K x_j\rangle}_{\text{(C) geom$\to$content}}
+
\underbrace{\langle E_q\rvec_i,\,E_k\tilde\rvec_j\rangle}_{\text{(D) geometry}},
\end{aligned}
\label{eq:attn-decomp}
\end{equation}
where $\tilde\rvec_j\!=\!(\mvec_j,\dvec_j)$ is the flipped coordinate of token $j$. Term~(A) is the original content attention. Terms~(B) and~(C) admit the same reading as~(D). A dot product against a projected ray is a linear functional on Pl\"ucker coordinates, and because the reciprocal product is a nondegenerate pairing, every such functional equals the reciprocal product with a fixed element of line space, a line or, in general, a screw~\cite{pottmann2001computational}; the flip makes this identification componentwise. 
Each cross term therefore defines a content-conditioned linear
functional over the ray at the other token. These mixed interactions
allow the compatibility between two tokens to depend jointly on
their learned representations and prescribed viewing rays, rather
than on independent content--content and ray--ray scores alone.
The terms enter as separate summands, so each can be switched off and its contribution measured (Section~\ref{sec:exp-component}). Term~(D) is the geometry-only interaction:
\begin{equation}
\text{(D)}= \rvec_i^\top M\,\tilde\rvec_j,\qquad
M = E_q^\top E_k\in\mathbb{R}^{6\times 6}.
\label{eq:term-D}
\end{equation}
\paragraph{Flip as a canonical basis.}
The role of the flip is best stated as a choice of basis on the geometric Q/K projection space rather than an algorithmic necessity:

\begin{proposition}
\label{prop:flip}
For the raw 6D construction, if
\(M=E_q^\top E_k=I_6\), term~(D) becomes
\[
  \text{(D)}
  =
  \dvec_i\!\cdot\!\mvec_j
  +
  \mvec_i\!\cdot\!\dvec_j
  =
  \plinner{\rvec_i}{\rvec_j}.
\]
\end{proposition}

\noindent(Proof in Appendix~\ref{sec:app-flip}.) For unconstrained
learnable $E_q,E_k$, the flipped and unflipped parameterizations
realize the same hypothesis class --- any $E_k$ in one form is
equivalent to $E_kP$ in the other, where $P$ swaps the
$(\dvec,\mvec)$ blocks (Appendix~\ref{sec:app-flip-reparam}). We do
not claim the flip enlarges the function space the model can express.
Its role is to choose a coordinate system in which the symmetric,
geometrically meaningful configuration $E_q=E_k=I_6$ corresponds to
the textbook Pl\"ucker reciprocal product, so that $M$ admits a
direct post-hoc reading: $\|M-I_6\|$ measures how far the learned
bilinear bias has drifted from the ray-incidence form. The empirical
gap between flipped and unflipped configurations is small
(Section~\ref{sec:exp-component}).

\paragraph{Parameter and compute cost.}
In the practical NGI version, each layer adds two $d\!\times\!7$ Q/K
projections, lightweight gates, and RMSNorms. The per-token projection
cost remains $O(N\!\cdot\!d)$, which is small compared with the
$O(N^2 d)$ cost of attention.

\subsection{Normalize-Gate-Inject (NGI)}
\label{sec:method-ngi}

The flip above defines where geometry enters attention. To use it on
mixed real-world data, we still need to control the scale of the
injected signal. This is where NGI is used.

The first issue is \textbf{scale heterogeneity}. The moment
$\mvec\!=\!\ovec\!\times\!\dvec$ scales linearly with the camera
translation magnitude $\|\ovec\|$. Different datasets use different
scale conventions, such as SfM-normalized scenes~\cite{schoenberger2016sfm}, DROID-SLAM~\cite{durrant2006simultaneous,teed2021droidslam} internal
scale, or metric meters. Even within one dataset, different baselines
can produce moments with very different magnitudes. A raw projection
$E\rvec$ would therefore mix a unit direction $\dvec$ with a
scale-dependent moment $\mvec$, making the geometry term unstable
across batches.

The second issue is \textbf{magnitude imbalance} with the content
branch.
Modern video DiTs apply QKNorm (RMS normalization) to $q$ and $k$
before the dot product, giving the content branch a controlled
magnitude. An unnormalized $E\rvec$ added on top does not respect
this normalization. A single global scalar is also too coarse: it
cannot handle both per-clip pose scale and the local balance between
geometry and content.

NGI handles these two issues in three steps:
\paragraph{Step 1: Direction/magnitude decomposition.}
We decouple the Pl\"ucker coordinate into a scale-invariant
directional part and a scalar log-magnitude:
\begin{equation}
  \hat{\mvec} = \mvec / \max(\|\mvec\|,\epsilon),\qquad
  s = \log\max(\|\mvec\|,\epsilon),
  \label{eq:decompose}
\end{equation}
where $\epsilon$ is a small numerical constant. The 7D geometric input
to $E_q$ is
$f^{(q)}_i\!=\!(\dvec_i,\hat{\mvec}_i,s_i)$, and to $E_k$ (after
the Q/K flip) is $f^{(k)}_j\!=\!(\hat{\mvec}_j,\dvec_j,s_j)$. The
first six dimensions are invariant to a global rescaling of camera
positions; 
the final dimension retains the log moment magnitude as an explicit pose-scale cue.

\paragraph{Step 2: Scale-aware gating.}
A small two-layer MLP $G_s$ maps the per-token log-magnitude to a
per-channel sigmoid gate:
\begin{equation}
  g_i = \sigma\bigl(G_s(s_i)\bigr) \in (0,1)^d,
  \label{eq:scale-gate}
\end{equation}
with biases initialized so that $g\!\approx\!0.5$ at $s\!=\!0$. 
The gate lets the model adjust the geometry injection according to
the encoded moment magnitude, rather than using one fixed strength
across all pose-scale conventions.

\paragraph{Step 3: Normalize and inject.}
We project the 7D feature, apply a learnable RMSNorm analogous to the
content-side QKNorm, and modulate the result by the gate:
\begin{equation}
  \mathrm{pe}^q_i = g_i\odot N_q\!\bigl(E_q\,f^{(q)}_i\bigr),\qquad
  \mathrm{pe}^k_j = g_j\odot N_k\!\bigl(E_k\,f^{(k)}_j\bigr),
  \label{eq:pe-rmsnorm}
\end{equation}
where $N_q,N_k$ are per-layer learnable RMSNorms. A scalar $\alpha$,
initialized to zero, controls the overall geometry/content balance:
\begin{align}
  q_i &= \mathrm{QKNorm}(\tilde W_Q x_i) + \alpha\cdot\mathrm{pe}^q_i,
  \label{eq:ngi-q}\\
  k_j &= \mathrm{QKNorm}(\tilde W_K x_j) + \alpha\cdot\mathrm{pe}^k_j.
  \label{eq:ngi-k}
\end{align}
Since both branches are normalized, $\alpha$ has a consistent meaning
across layers and training examples.

\begin{algorithm}[ht]
\caption{One forward-backward step of \method on top of a pretrained video DiT.}
\label{alg:step}
\begin{algorithmic}[1]
\Require Latent video $z_0$, text embedding $c$, per-frame extrinsics
  $\{T_t\}$ and intrinsics $\{K_t\}$, dataset id $\mathrm{ds}$, near-depth $z_n$ (or $\bot$).
\State Apply trajectory rescaling: $\ovec_t \gets (\eta_{\mathrm{ds}}/\max(z_n,\epsilon))\cdot\ovec_t$
       \Comment{Appendix~\ref{sec:training-trajscale}--\ref{sec:training-neardepth}}
\State Compute per-token rays $\{(\dvec_i,\mvec_i)\}$ from
       $\{T_t,K_t\}$ and the latent patch grid
       \Comment{Section~\ref{sec:bg-camera}}
\State Sample noise $\epsilon\sim\mathcal{N}(0,I)$, timestep $\tau$, build noisy latent $z_\tau$
\For{each DiT block $\ell=1\dots L$}
    \State $q,k,v\gets W_Q^\ell h,\,W_K^\ell h,\,W_V^\ell h$ \Comment{Pretrained projections, possibly fine-tuned}
    \State $q,k\gets \mathrm{QKNorm}(q),\,\mathrm{QKNorm}(k)$
    \State $q,k\gets R(u,v,t)\cdot q,\,R(u,v,t)\cdot k$ \Comment{Standard 3D RoPE}
    \State $f^{(q)}\gets(\dvec,\hat{\mvec},s)$, $f^{(k)}\gets(\hat{\mvec},\dvec,s)$ where $s=\log\|\mvec\|$
    \State $g\gets\sigma(G_s^\ell(\tilde s))$ \Comment{$\tilde s$ optionally aug.\ per Eq.~\eqref{eq:logscale-aug}}
    \State $q\gets q + \alpha\cdot g\odot N_q^\ell(E_q^\ell f^{(q)})$
    \State $k\gets k + \alpha\cdot g\odot N_k^\ell(E_k^\ell f^{(k)})$
    \State $h\gets h + \mathrm{Attn}(q,k,v)\,W_O^\ell$
    \State $h\gets h + \mathrm{CrossAttn}(h,c) + \mathrm{FFN}(h)$
\EndFor
\State Compute diffusion / flow loss between predicted noise/velocity and target and backpropagate.
\end{algorithmic}
\end{algorithm}
\vspace{-1.5em}

\paragraph{Log-scale augmentation.}
We also apply a simple scale augmentation during training. When
computing the gate, we optionally perturb $s$ by a clip-level offset:
\begin{equation}
  \tilde s_i = s_i + \delta,\quad
  \delta\sim\mathrm{Uniform}(-1.2,\,1.6),\;\text{with prob.\ }0.3,
  \label{eq:logscale-aug}
\end{equation}
shared across all tokens of a clip. This simulates a global rescaling
$\ovec\!\to\!e^\delta\ovec$ of the camera trajectory without changing
the supervision. Only the gate receives $\tilde s_i$; the projection
$E_q f^{(q)}$ still uses the true $s_i$.

\subsection{Architectural Discussion}
\label{sec:method-impl}

\paragraph{Pseudocode.}
\label{sec:training-pseudo}
The full one-step procedure is given in Algorithm~\ref{alg:step}.
The per-token Pl\"ucker computation, decomposition
$(\dvec,\hat{\mvec},s)$, scale gate, and Q/K flip are applied
independently in each attention layer.

\paragraph{Relation to alternative camera-aware positional encodings.}
\label{sec:method-discussion}
Unlike camera-aware positional encodings that replace or partition
RoPE~\cite{li2025prope,zhang2026ucpe,li2026rerope}, \method preserves
the pretrained 3D RoPE and injects geometric information as an
additional Q/K bias term. Quantitative comparisons with these
alternatives are provided in
Section~\ref{sec:experiments}.

%% file: sections/06_experiments.tex
\section{Experiments}
\label{sec:experiments}

We evaluate \method along three axes: (i) video generation quality, (ii) camera-trajectory controllability, and (iii) cross-frame geometric consistency. We compare \method with four camera-conditioned video generation baselines and four alternative camera-aware positional encoding designs on the same backbone, isolating the contribution of the proposed Pl\"ucker-flip/NGI formulation.

\subsection{Setup}
\label{sec:exp-setup}

\paragraph{Backbone.}
For the main comparisons (Table~\ref{tab:main}), we evaluate our method on two official variants from Wan-2.2~\cite{wan2025wan}: the 5B TI2V model and the 14B I2V model. To ensure computational efficiency, all ablation studies are conducted exclusively using the Wan-2.2-TI2V-5B model. Across all runs, videos are generated at $480\!\times\!832$ resolution with $T\!=\!81$ frames ($T_z\!=\!21$ latent frames), and the models are trained on the four-dataset mixture (Appendix~\ref{sec:training}). All variants share their respective pretrained weights; only the newly integrated camera-conditioning module differs.

\paragraph{Baselines.}
We compare against five external methods: CameraCtrl~\cite{he2025cameractrl} (Pl\"ucker maps via ControlNet encoder), ReCamMaster~\cite{bai2025recammaster} (video-to-video re-rendering with camera control),
ProPE~\cite{li2025prope} (deployed via a zero-initialized parallel branch), UCPE~\cite{zhang2026ucpe} (full camera geometry via spatial-attention adapter), and ReRoPE~\cite{li2026rerope} (low-frequency temporal RoPE replaced by relative camera pose). Where official checkpoints are unavailable we re-implement on our backbone under the same data and budget (Appendix~\ref{sec:appendix}).
We additionally train four \emph{FreqSplit-RoPE} variants that replace the low-frequency half of temporal RoPE \emph{multiplicatively} with (i)~a $4\!\times\!4$ projective transform, (ii)~camera-plane near/far UV coordinates, (iii)~learned Pl\"ucker-to-phase mapping, or (iv)~analytical $(\theta,\phi,u,v)$ phases. All four share our backbone and budget; \method differs by adding Pl\"ucker features \emph{additively} without modifying RoPE.

\vspace{-1em}
\paragraph{Evaluation.}
We evaluate on a held-out RE10K~\cite{zhou2018re10k} split (image-to-video: first GT frame + GT trajectory) and report eight metrics.
\emph{Quality:} \textbf{CLIP}$\uparrow$ (inter-frame similarity).
\emph{Camera controllability:} \textbf{RotErr}$\downarrow$ (geodesic rotation error), \textbf{TransErr}$\downarrow$ (L2 translation error), \textbf{CamMC}$\downarrow$ (per-frame pose Frobenius norm), \textbf{ATE}$\downarrow$ (RMS absolute trajectory error), all on per-frame poses anchored to frame~0.
\emph{Distribution:} \textbf{FVD}$\downarrow$ / \textbf{FVD$_{\mathrm{c}}$}$\downarrow$ (full-frame / center-cropped Fr\'echet Video Distance~\cite{unterthiner2019fvd}), \textbf{FID}$\downarrow$ (per-frame Fr\'echet Inception Distance).
Trajectories are estimated from both GT and generated videos using ViPE~\cite{huang2025vipe} and compared \emph{without rescaling}. Unlike the common protocol of normalizing trajectories by their maximum translation magnitude, we preserve absolute-scale fidelity. This avoids artificially inflating the apparent controllability of methods that produce only small motions. Results under the conventional rescaled protocol are provided in the Appendix; the ranking is unchanged.
All methods use 50 denoising steps with CFG scale 5.0 (text) / 1.0 (camera), matching Appendix~\ref{sec:training-opt}.

\begin{figure*}[h]
  \centering
  \includegraphics[width=0.95\textwidth]{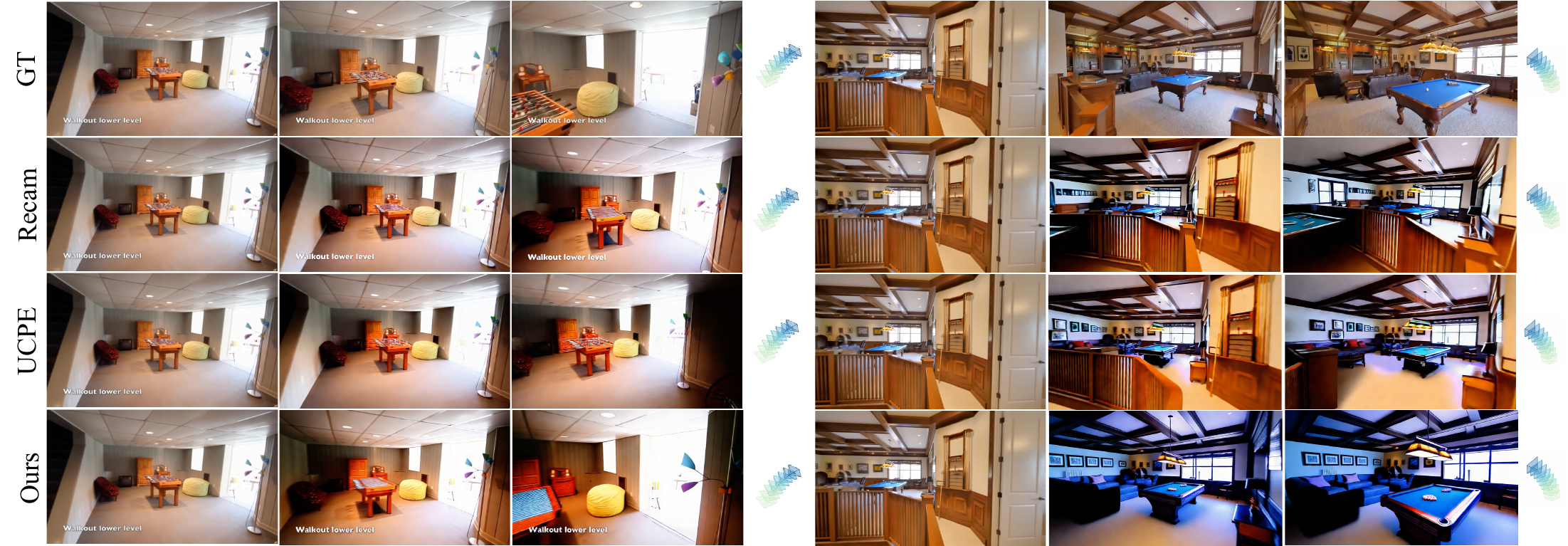}
  \vspace{-0.35em}
  \caption{Qualitative comparison on RealEstate10K. Each row shows
    frames from the same clip generated under the same target camera
    trajectory. Top to bottom: ground truth,
    ReCamMaster~\cite{bai2025recammaster},
    UCPE~\cite{zhang2026ucpe}, and \method (ours). \method follows the
    target trajectory more faithfully across frames; ReCamMaster and
    UCPE drift from the camera path.}
  \Description{A four-row grid of video frames sampled at uniform
    intervals along a RealEstate10K camera trajectory. The top row is
    the ground-truth video; the next three rows show ReCamMaster,
    UCPE and our method, all generated from the same first frame and
    camera-pose conditioning.}
  \label{fig:re10k-qual}
\end{figure*}

\begin{table*}[h]
  \centering
\caption{Main comparison on the RE10K held-out split across different backbone scales. Higher is
  better for CLIP; lower is better for the rest. Pose metrics
  (RotErr, TransErr, CamMC, ATE) are computed on raw, unscaled
  ViPE trajectories estimated identically on both GT and generated
  videos (Section~\ref{sec:exp-setup});
  rescaled-alignment results matching the conventional protocol
  are in Appendix~\ref{sec:appendix}.}
  \label{tab:main}
  \scriptsize
  \resizebox{0.8\textwidth}{!}{%
  \begin{tabular}{l c c c c c c c c}
    \toprule
    & \multicolumn{1}{c}{Quality} & \multicolumn{4}{c}{Camera controllability} & \multicolumn{3}{c}{Distribution} \\
    \cmidrule(lr){2-2}\cmidrule(lr){3-6}\cmidrule(lr){7-9}
    Method
      & CLIP$\uparrow$
      & RotErr$\downarrow$ & TransErr$\downarrow$ & CamMC$\downarrow$ & ATE$\downarrow$
      & FVD$\downarrow$ & FVD$_{\mathrm{c}}\downarrow$ & FID$\downarrow$ \\
    \midrule
    \multicolumn{9}{@{}l}{\textit{Wan-2.2 5B Scale}}             \\
    CameraCtrl~\cite{he2025cameractrl}        & 25.04  & 0.152 & 1.292 & 1.355 & 1.501 & 824.37 & 805.62 & 64.08 \\
    ProPE~\cite{li2025prope}                  & 24.90  & 0.117 & 1.653 & 1.694 & 1.915 & 717.2 & 611.5   & 58.33 \\
    ReCamMaster~\cite{bai2025recammaster}     & 24.97  & 0.131 & 1.226 & 1.279 & 1.460 & 874.30 & 890.52 & 62.53 \\
    ReRoPE~\cite{li2026rerope}                & 25.20  & 0.137 & 1.109 & 1.215 & 1.350 & 684.57 & 650.31 & 60.77 \\
    UCPE~\cite{zhang2026ucpe}                 & 25.39  & 0.113 & 0.856 & 0.909 & 0.990 & 703.41 & 755.83 & 61.50 \\
    \method (ours)                            & \textbf{26.05}  & \textbf{0.085} & \textbf{0.751} & \textbf{0.802} & \textbf{0.884} & \textbf{543.17} & \textbf{588.62} & \textbf{57.83} \\
    \midrule
    \multicolumn{9}{@{}l}{\textit{Wan-2.2 14B Scale}} \\
    ReCamMaster~\cite{bai2025recammaster}     & 25.31      & 0.109 & 0.976 & 1.012 & 0.995 & 675.23  & 697.10 & 59.21     \\ 
    ReRoPE~\cite{li2026rerope}                & 25.85  & 0.114 & 0.820 & 0.905 & 0.859 & 493.30 & 525.61 & 49.18 \\
    UCPE~\cite{zhang2026ucpe}                 & 25.72  & 0.082 & 0.693 & 0.760 & 0.788 & 529.42 & 558.70 & 54.75 \\
    \method (ours)                            & \textbf{26.30}  & \textbf{0.058} & \textbf{0.517} & \textbf{0.530} & \textbf{0.605} & \textbf{280.17} & \textbf{354.52} & \textbf{41.01} \\
    \bottomrule
  \end{tabular}}
\end{table*}

\subsection{Main Comparison}
\label{sec:exp-main}

\begin{figure*}[!ht]
  \centering
  \includegraphics[width=0.95\textwidth]{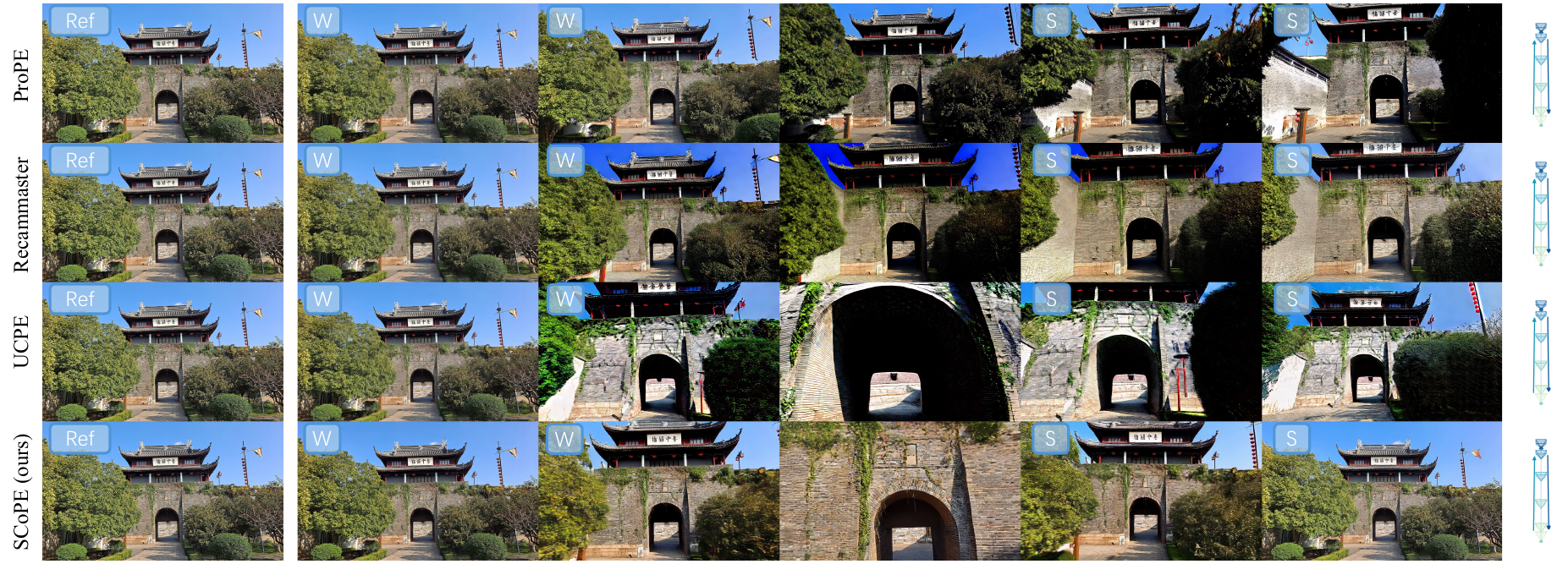}
  \vspace{-0.35em}
  \caption{\textbf{Closed-loop revisit comparison.} Starting from the same reference image (Ref), each method follows the same camera loop, moving away from and then returning to the initial view. Right: input camera trajectory of each generated video.}
  \Description{A five-row grid of video frames generated by our
    method on different first-frame scenes, each driven by its own
    camera trajectory shown at the right of the row.}
  \label{fig:revisit_compare}
\end{figure*}
\begin{figure*}[!ht]
  \centering
  \includegraphics[width=0.95\textwidth]{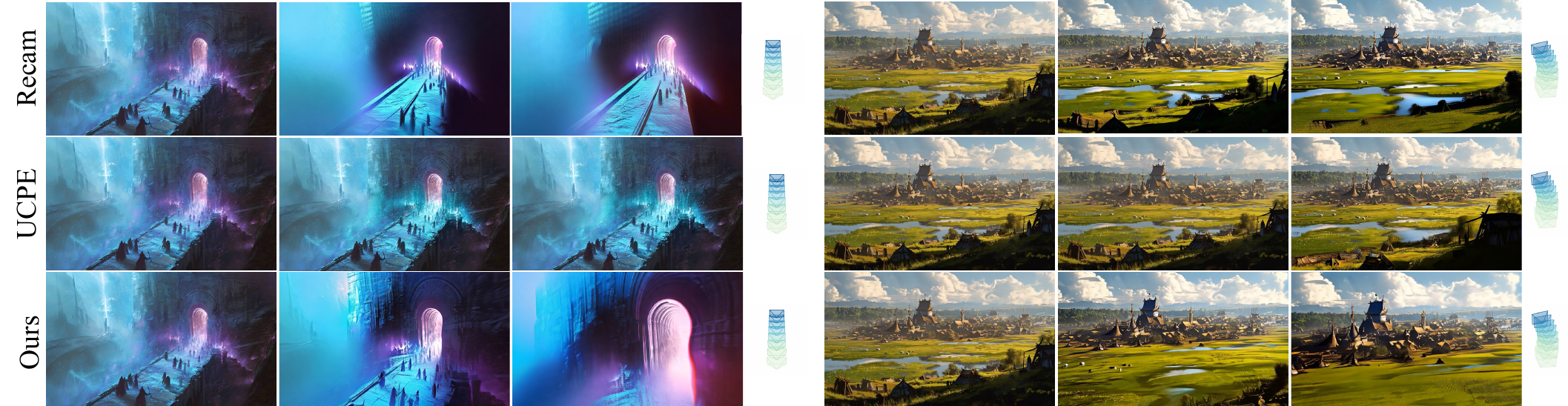}
  \vspace{-0.35em}
  \caption{Out-of-distribution qualitative comparison. Inputs are
    artistic paintings (rather than natural photographs), conditioned
    on simple camera trajectories such as a forward dolly. No
    ground-truth video exists in this setting, so each row shows only
    the generated frames of a single method. \method realizes the
    intended camera motion (e.g., zoom-in with correct foreground/
    background parallax) on stylized inputs, while
    ReCamMaster~\cite{bai2025recammaster} and
    UCPE~\cite{zhang2026ucpe} fail to follow the trajectory.}
  \Description{Out-of-distribution qualitative comparison figure with
    three rows (ReCamMaster, UCPE, Ours), each row showing generated
    frames from an artistic painting first frame under a prescribed
    camera trajectory.}
  \label{fig:qual-ood}
\end{figure*}
\begin{figure*}[!ht]
  \centering
  \includegraphics[width=0.98\textwidth]{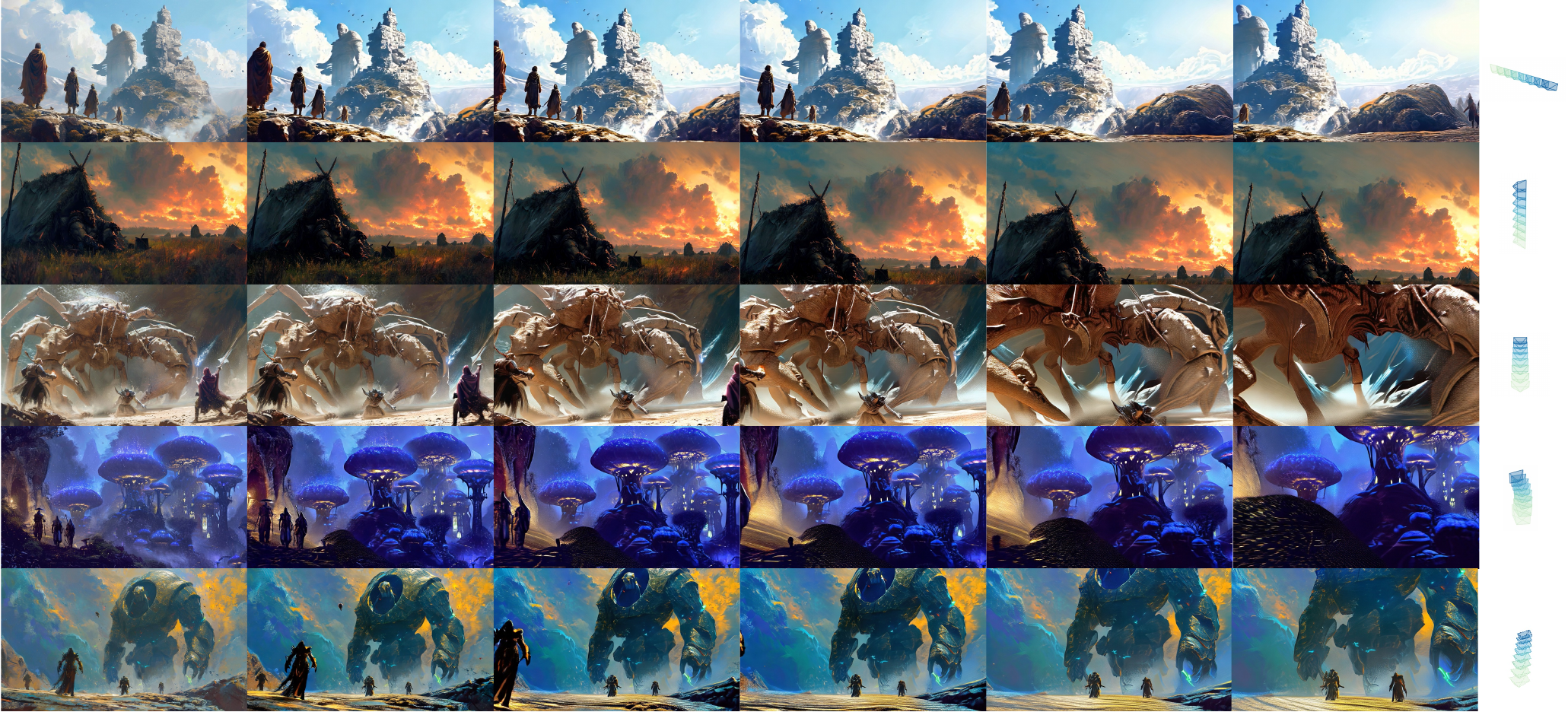}
  \vspace{-0.35em}
  \caption{Generalization to stylized inputs. Each row is a
    different hand-painted concept-art first frame driven by a
    distinct target camera trajectory with \method; frames go left
    to right along time, and the rightmost column shows the input
    trajectory as a stack of camera frusta.}
  \Description{A five-row grid of video frames generated by our
    method on hand-painted concept-art first frames, each driven
    by its own camera trajectory shown at the right of the row.}
  \label{fig:qual-style-gallery}
\end{figure*}
\begin{figure*}[h]
  \centering
  \includegraphics[width=0.98\textwidth]{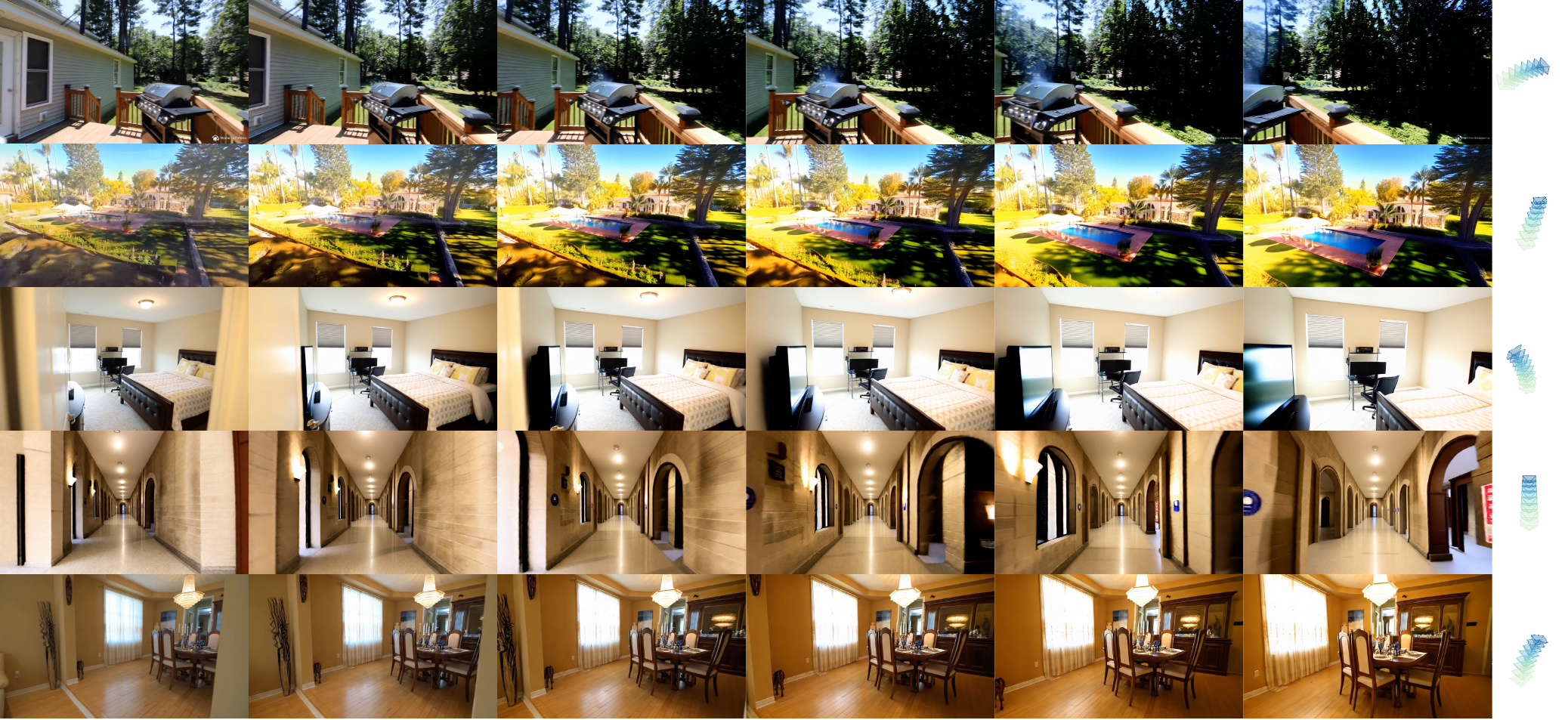}
  \vspace{-0.35em}
  \caption{Gallery of \method on diverse scenes. Each row is a
    different first frame driven by a distinct target camera
    trajectory; frames go left to right along time, and the
    rightmost column shows the input trajectory as a stack of
    camera frusta.}
  \Description{A five-row grid of video frames generated by our
    method on different first-frame scenes, each driven by its own
    camera trajectory shown at the right of the row.}
  \label{fig:qual-rel-gallery}
\end{figure*}
\begin{figure*}[!ht]
  \centering
  \includegraphics[width=\textwidth]{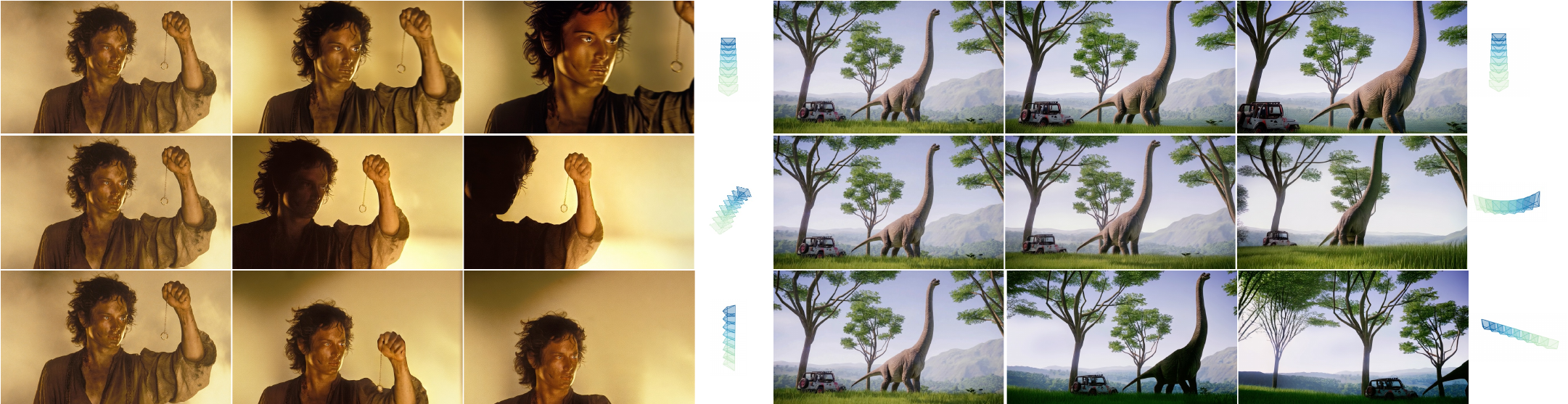}
  \vspace{-0.35em}
  \caption{Multi-trajectory gallery on cinematic out-of-distribution
    inputs. We fix two movie-still first frames and drive each
    through several distinct camera trajectories (one trajectory per
    row within a scene), all generated by \method.
 }
  \Description{Trajectory gallery on two movie-still first frames.
    Each scene is shown under several camera trajectories generated
    by our method, with sampled frames laid out from left to right
    along time.}
  \label{fig:qual-movie}
\end{figure*}

Table~\ref{tab:main} reports the main quantitative comparison and
Figures~\ref{fig:re10k-qual}--\ref{fig:qual-movie} show qualitative
results. Consistent with the design motivation in
Sections~\ref{sec:method-decomp} and~\ref{sec:method-ngi}, \method
improves all four pose-error metrics substantially while also
improving CLIP, FVD, FVD$_{\mathrm{c}}$, and FID. We organize the qualitative evaluation along three progressively
harder axes. \emph{In-domain comparison}
(Figure~\ref{fig:re10k-qual}): on the RE10K validation split, we
show the same clip generated by ReCamMaster, UCPE, and \method under
identical camera conditioning; \method follows the target trajectory
more faithfully across frames while the baselines drift.
\emph{In-domain gallery}
(Figure~\ref{fig:qual-rel-gallery}): a broader set of RE10K scenes
where each row is a different first frame driven by a distinct
trajectory, illustrating consistent camera following across indoor
walkthroughs, outdoor pans, and forward-moving corridors.
\emph{Out-of-distribution generalization}
(Figures~\ref{fig:qual-ood},~\ref{fig:qual-style-gallery},
and~\ref{fig:qual-movie}): we test on inputs that never appear in
the training data. Figure~\ref{fig:qual-ood} compares baselines on
artistic paintings, where \method realizes the intended camera motion
with correct parallax while baselines fail to follow the trajectory.
Figure~\ref{fig:qual-style-gallery} extends this to a wider set of
hand-painted stylized first frames, showing that \method preserves
artistic style while accurately executing the prescribed camera
path. Figure~\ref{fig:qual-movie} fixes two cinematic movie-still
first frames and drives each through several distinct trajectories,
demonstrating that the geometric encoding generalizes across both
content domain and trajectory diversity.
Additional evaluations of dynamic-scene quality and inference
efficiency are provided in
Appendix~\ref{sec:app-vbench} and~\ref{sec:app-efficiency},
respectively.

\begin{table*}[!ht]
\centering
\caption{Revisit evaluation on the test set. Pose metrics
measure whether the generated camera returns to its initial pose,
while appearance metrics measure recovery of the conditioning view.
}
\label{tab:revisit}
\scriptsize
\resizebox{0.9\textwidth}{!}{%
\begin{tabular}{lcccccc}
\toprule
& \multicolumn{3}{c}{Camera return}
& \multicolumn{3}{c}{Appearance recovery} \\
\cmidrule(lr){2-4}\cmidrule(lr){5-7}
Method
& Return Trans.\ Ratio $\downarrow$
& Return Rot.\ ($^\circ$) $\downarrow$
& Norm.\ ATE $\downarrow$
& Return LPIPS $\downarrow$
& LPIPS Recovery $\uparrow$
& DINOv2 Sim.\ $\uparrow$ \\
\midrule
ProPE~\cite{li2025prope}
& 0.784 & 4.90 & 0.533 & 0.603 & $-0.176$ & 0.874 \\
ReCamMaster~\cite{bai2025recammaster}
& 0.859 & 3.80 & 0.625 & 0.653 & 0.094 & 0.890 \\
UCPE~\cite{zhang2026ucpe}
& 0.433 & 3.42 & 0.391 & 0.403 & 0.102 & 0.910 \\
\method (ours)
& \textbf{0.090}
& \textbf{1.89}
& \textbf{0.275}
& \textbf{0.230}
& \textbf{0.587}
& \textbf{0.952} \\
\bottomrule
\end{tabular}}
\end{table*}

\subsection{Closed-Loop Revisit Consistency}

To evaluate long-horizon closed-loop consistency and view recovery, we
introduce a Revisit protocol in which each method starts from the same
conditioning image, moves away from the initial view, and finally
returns to the starting pose. For each conditioning image, we generate
four closed-loop camera paths: truck, orbit, dolly, and a
non-palindromic loop. We evaluate an away window near the maximum
camera excursion (frames 36--40) and a return window after the loop is
completed (frames 76--80). Camera trajectories are estimated with
ViPE~\cite{huang2025vipe} and expressed relative to frame~0.

\textbf{Return Translation Ratio} is the mean camera-center
displacement from frame~0 over the return window, divided by the
maximum displacement over the full generated trajectory.
\textbf{Return Rotation} is the mean geodesic rotation from the
initial orientation over the return window. \textbf{Normalized ATE}
compares the predicted and commanded trajectories after anchoring
each to frame~0 and independently normalizing it by its maximum
translation magnitude. For appearance consistency, \textbf{Return
LPIPS} and \textbf{DINOv2 Similarity} compare the conditioning image
with the frames in the return window, using AlexNet LPIPS and DINOv2
global-feature cosine similarity, respectively. We further define
$\textbf{LPIPS Recovery}=1-\mathrm{LPIPS}_{\mathrm{return}}/
\mathrm{LPIPS}_{\mathrm{away}}$, which measures the fraction of the
appearance discrepancy at maximum excursion that is recovered after
returning. Window-based scores are first averaged over the
corresponding five frames, then over the four trajectories of each
conditioning image, and finally across cases. Appearance metrics are
interpreted jointly with the pose metrics, since similarity to the
conditioning image alone does not demonstrate that the prescribed
camera motion was executed.

As shown in Table~\ref{tab:revisit}, \method ranks first on all six
metrics. Relative to the strongest baseline for each metric, it
reduces Return Translation Ratio by $79\%$, Return Rotation by $45\%$,
Normalized ATE by $30\%$, and Return LPIPS by $43\%$. It also raises
LPIPS Recovery from $0.102$ to $0.587$ and DINOv2 Similarity from
$0.910$ to $0.952$. The simultaneous gains in pose return and
revisited-view fidelity indicate that \method recovers the initial
view after executing the commanded excursion, rather than preserving
appearance by suppressing camera motion.

Figure~\ref{fig:revisit_compare} illustrates the resulting behavior: all
methods are driven away from the same conditioning view by the same
closed-loop command. ProPE and ReCamMaster accumulate substantial
return error, while UCPE only partially restores the initial view.
In contrast, \method returns both the camera and the generated content
closest to their initial states.

\subsection{Ablations}
\label{sec:exp-abl}

We organize ablations into two complementary tables. The first
contrasts \method against four alternative camera-aware PE designs
that share our backbone (Section~\ref{sec:exp-pe-design}). The second
removes one component at a time from \method, isolating the
contribution of each design element (Section~\ref{sec:exp-component}).

\begin{table*}[t]
  \centering
  \caption{Camera-aware PE design space on RE10K. All rows share the
    Wan2.2-TI2V-5B backbone, the four-dataset training mixture and
    the same training budget; the only difference is how the camera
    enters attention. \method adds an \emph{additive} 6D Pl\"ucker
    term to $q,k$, while the four FreqSplit-RoPE variants
    \emph{multiplicatively replace} the low-frequency half of the
    temporal RoPE.}
  \label{tab:abl-pe-design}
  \scriptsize
  \resizebox{0.8\textwidth}{!}{%
  \begin{tabular}{l c c c c c c c c}
    \toprule
    & \multicolumn{1}{c}{Quality} & \multicolumn{4}{c}{Camera controllability} & \multicolumn{3}{c}{Distribution} \\
    \cmidrule(lr){2-2}\cmidrule(lr){3-6}\cmidrule(lr){7-9}
    PE design
      & CLIP$\uparrow$
      & RotErr$\downarrow$ & TransErr$\downarrow$ & CamMC$\downarrow$ & ATE$\downarrow$
      & FVD$\downarrow$ & FVD$_{\mathrm{c}}\downarrow$ & FID$\downarrow$ \\
    \midrule
    FreqSplit-RoPE + 4$\times$4 Proj.        & 25.09  & 0.175 & 1.227 & 1.308 & 1.584 & 675.03 & 723.90 & 62.98 \\
    FreqSplit-RoPE + Camera-UV               & 25.39 & 0.141 & 1.205 & 1.394 & 1.602 & 685.20 & 720.11 & 60.74 \\
    FreqSplit-RoPE + Pl\"ucker-RoPE          & 25.61 & 0.159 & 1.018 & 1.323 & 1.447 & 671.45 & 659.03 & 59.31 \\
    FreqSplit-RoPE + 4DOF Pl\"ucker          & 25.52 & 0.137 & 0.976 & 1.228 & 1.419 & 638.31 & 645.29 & 58.54 \\
    \midrule
    \method (ours)                           & \textbf{26.05} & \textbf{0.085} & \textbf{0.751} & \textbf{0.802} & \textbf{0.884} & \textbf{543.17} & \textbf{588.62} & \textbf{57.83} \\
    \bottomrule
  \end{tabular}}
\end{table*}

\subsubsection{Camera-Aware PE Design Space}
\label{sec:exp-pe-design}


A direct multiplicative camera PE would replace the pretrained RoPE
across all frequency channels. In our experiments, this
full-frequency replacement failed to yield a usable
camera-controlled model under the matched fine-tuning budget,
suggesting disruption of the pretrained positional prior. We
therefore follow the frequency-splitting strategy of
ReRoPE~\cite{li2026rerope}: each alternative retains the
high-frequency half of temporal RoPE and replaces its low-frequency
half with a camera-derived transformation. This partial replacement
provides a stable common scaffold for comparing different
multiplicative camera encodings.

Within this common FreqSplit interface, we test four ways of
representing camera geometry: a $4\times4$ projective transform,
near/far camera-plane coordinates, learned rotary phases from the full
6D Pl\"ucker ray, and analytical phases from its four intrinsic
degrees of freedom. All variants share the same backbone, data,
training budget, and frequency allocation. This controlled design
study asks whether a stable multiplicative retrofit, equipped with
either projective or ray-level geometry, can recover the benefit of
\method. The Pl\"ucker-RoPE variant provides the closest comparison:
it receives the same full ray coordinate as \method, but maps that
coordinate to multiplicative rotary phases in the replaced frequency
bands.

\method takes a different retrofit route. It preserves every channel
of the pretrained RoPE and adds a zero-initialized Pl\"ucker term to
$q$ and $k$. It therefore starts from the exact pretrained attention
operator while introducing the content--geometry cross terms and
geometry-only pairing derived in
Section~\ref{sec:method-decomp}. As shown in
Table~\ref{tab:abl-pe-design}, \method outperforms all four viable
multiplicative alternatives on every metric. The gap to
Pl\"ucker-RoPE is especially informative: using the same ray
coordinate through rotary phases is not sufficient to reproduce the
gain of the additive, RoPE-preserving formulation.

\subsubsection{Component Ablation}
\label{sec:exp-component}

Table~\ref{tab:abl-component} reports component ablations of the full \method configuration. Each row removes a single component while keeping all other settings and training recipes unchanged.

\begin{table}[t]
  \centering
  \caption{Component ablation of \method on RE10K. ``Full'' is the
    configuration used for the main comparison. Each subsequent row
    removes a single component while keeping the rest unchanged.}
  \label{tab:abl-component}
  \footnotesize
  \setlength{\tabcolsep}{3.4pt}
  \renewcommand{\arraystretch}{1.08}
  \resizebox{\linewidth}{!}{\begin{tabular}{@{}l c c c c c@{}}
    \toprule
    Variant & CLIP$\uparrow$ & RotErr$\downarrow$ & CamMC$\downarrow$ & ATE$\downarrow$ & FVD$\downarrow$ \\
    \midrule
    Full \method                                          & \textbf{26.05} & \textbf{0.085} & \textbf{0.802} & \textbf{0.884} & \textbf{543.17} \\
    \midrule
    w/o Q/K flip                                    & 25.93 & 0.091 & 0.817 & 0.929 & 579.83 \\
    w/o \ngi                     & 25.86 & 0.086 & 0.865 & 0.933 & 560.37 \\
    w/o PE RMSNorm                                  & 25.62 & 0.090 & 0.874 & 0.913 & 601.39 \\
    w/o log-scale aug.                              & 25.94 & 0.083 & 0.823 & 0.890 & 557.92 \\
    w/o zero init                                   & 25.68 & 0.086 & 0.837 & 0.899 & 635.42 \\
    with V transform                                & 25.57 & 0.090 & 0.855 & 0.949 & 724.10 \\
    w/o (B)+(C) content$\leftrightarrow$geometry  & 25.21 & 0.135 & 1.174 & 1.358 & 695.41 \\
    w/o (D) geometry$\leftrightarrow$geometry     & 24.08 & 0.159 & 1.382 & 1.440 & 733.08 \\
    \bottomrule
  \end{tabular}}
\end{table}

\paragraph{What each row tests.}
The \textbf{w/o Q/K flip} row sends $(\dvec,\mvec)$ on both sides
instead of the flipped pair; under arbitrary learnable $E_q,E_k$ the
flip is a reparametrization
(Appendix~\ref{sec:app-flip-reparam}), so only a marginal change is
expected, which the row confirms.
\textbf{w/o \ngi} reduces the encoding to a raw 6D Pl\"ucker
projection $E\rvec$, removing direction/magnitude decomposition,
scale gating, and PE RMSNorm together --- the largest aggregate
ablation of the normalize-gate-inject pipeline.
\textbf{w/o PE RMSNorm} keeps decomposition and gating but drops
the PE-side RMSNorm, so the geometry magnitude is no longer matched
to the content branch and a single $\alpha$ cannot balance both.
\textbf{w/o log-scale aug.} removes random log-magnitude
augmentation during training, isolating the role of synthetic
scale diversity in stabilizing cross-domain generalization; the
modest drop suggests the augmentation helps but is not the
dominant ingredient.
\textbf{w/o zero init} initializes $E_q,E_k$ randomly instead of
to identity (zero residual at $t\!=\!0$), forcing the model to
recover from a non-trivial perturbation of the pretrained
attention; the larger FVD drop shows that breaking the pretrained
equilibrium at the start of training is costly even when the final
solution is ultimately reachable.
\textbf{with V transform} additionally injects geometry into $v$,
which consistently underperforms Q/K-only injection and confirms
that ray geometry is most useful as an attention-bias positional
signal rather than as a value-channel feature.
The last two rows directly intervene on the attention decomposition
of Eq.~\eqref{eq:attn-decomp}: \textbf{w/o (B)+(C)
content$\leftrightarrow$geometry} keeps the content--content (A)
and geometry--geometry (D) terms but suppresses the two cross-modal
terms that couple appearance with ray geometry, while
\textbf{w/o (D) geometry$\leftrightarrow$geometry} removes the
pure geometric prior. Both produce sharp drops on every metric.

\vspace{-1em}


\paragraph{Complementary roles of mixed and geometric terms.}


The final two ablations separate the pure ray--ray interaction from
the mixed feature--ray interactions in
Eq.~\eqref{eq:attn-decomp}. Term~(A) is the standard
content--content attention term, while term~(D) evaluates the two
camera rays. Their sum has the separable form
\[
    A(x_i,x_j)+D(\rvec_i,\rvec_j),
\]
where visual affinity and camera compatibility are computed by two
independent pairwise functions. For a fixed query ray, the relative
preference induced by (D) over candidate rays is therefore
independent of the query feature.

Terms~(B) and~(C) introduce the mixed dependence between token
features and camera rays. Let \(c_i=\tilde W_Qx_i\). The part of the
attention logit that depends on the candidate ray can be written as
\[
  \text{(B)}+\text{(D)}
  =
  \left(E_k^\top c_i + M^\top\rvec_i\right)^\top
  \tilde\rvec_j .
\]
Here, \(M^\top\rvec_i\) gives the ray-dependent coefficient from
term~(D), while \(E_k^\top c_i\) adds a coefficient determined by
the query feature. The geometry-related preference over candidate
rays can consequently vary with the current query representation,
instead of being determined by camera geometry alone. Term~(C)
provides the complementary interaction, allowing the relevance of a
key feature to depend directly on the query ray. Together, (B) and
(C) remove the separability between visual content and camera
geometry, allowing the geometry-involving contribution to attention
to be conditioned on the current token representations.

The ablation supports these complementary roles. Removing (B)+(C)
degrades RotErr from $0.085$ to $0.135$, CamMC from $0.802$ to
$1.174$, ATE from $0.884$ to $1.358$, and FVD from $543.17$ to
$695.41$. Thus, a scene-independent ray prior combined with the
original content score does not recover the performance of their
explicit mixed interactions. Removing (D) causes a further
degradation, confirming that the ray--ray prior remains the primary
geometric component. The full model benefits from both: (D) supplies
a shared camera-geometric compatibility, while (B) and (C) adapt its
use to the representations formed for the current video.

\subsubsection{Data composition}
Table~\ref{tab:abl-data} ablates the effect of training-data composition. We compare training on RE10K alone, RE10K+DL3DV, RE10K+DL3DV+PanShot, and the full RE10K+DL3DV+PanShot+OmniWorld mixture. 
This highlights the value of \emph{normalize-gate-inject} as more scale-heterogeneous data is added: while methods lacking scale-invariance may degrade as the trajectory-scale distribution widens, \method consistently benefits from the additional data.

\begin{table}[t]
  \centering
  \caption{Effect of training-data composition on RE10K-test metrics.}
  \label{tab:abl-data}
  \footnotesize
  \setlength{\tabcolsep}{3.4pt}
  \renewcommand{\arraystretch}{1.08}
  \resizebox{\linewidth}{!}{%
  \begin{tabular}{@{}l c c c c c@{}}
    \toprule
    Training mixture & CLIP$\uparrow$ & RotErr$\downarrow$ & CamMC$\downarrow$ & ATE$\downarrow$ & FVD$\downarrow$ \\
    \midrule
    RE10K only                     & 25.58 & 0.091 & 0.897 & 0.983 & 586.45 \\
    RE10K + DL3DV                  & 25.65 & 0.089 & 0.872 & 0.960 & 602.13 \\
    + PanShot                      & 25.93 & \textbf{0.083} & 0.836 & 0.905 & 550.19 \\
    + OmniWorld (full)             & \textbf{26.05} & 0.085 & \textbf{0.802} & \textbf{0.884} & \textbf{543.17} \\
    \bottomrule
  \end{tabular}}
\end{table}

%% file: sections/08_conclusion.tex
\section{Conclusion}
\label{sec:conclusion}

We presented \method, a sightline-coordinate positional encoding for video
diffusion transformers that exploits the algebraic match between the
Pl\"ucker reciprocal product and the attention dot product. By adding
per-token Pl\"ucker coordinates to queries and keys with a flip
arrangement, \method makes the attention score contain a
whose canonical configuration recovers the Klein form,
giving the model a built-in 3D inductive bias before any learning.
Normalize-Gate-Inject further decouples ray direction from
translation magnitude and gates the encoding adaptively, making it
stable across datasets with heterogeneous pose scales. The full module
adds less than $0.1\%$ parameters to a pretrained 5B video DiT,
preserves the original RoPE, and requires no adapter or architectural
change. Quantitative and qualitative results show that \method improves camera controllability and cross-view and closed-loop consistency while preserving the generation quality of the pretrained backbone.

%% file: sections/A_appendix.tex
\section*{\LARGE Appendix}
\section{Additional Derivations}
\label{sec:appendix}

\begin{table*}[htbp]
  \centering
  \caption{Main comparison on the RE10K held-out split under the
    \emph{rescaled-alignment} protocol used in prior
    camera-controllability work. Predicted and ground-truth ViPE
    trajectories are normalized to a common translation norm before
    pose metrics (RotErr, TransErr, CamMC, ATE) are computed.
    Everything else matches Table~\ref{tab:main}. Higher is better
    for CLIP; lower is better for the rest.}
  \label{tab:main-rescaled}
  \small
  \begin{tabular}{l c c c c c c c c}
    \toprule
    & \multicolumn{1}{c}{Quality} & \multicolumn{4}{c}{Camera controllability} & \multicolumn{3}{c}{Distribution} \\
    \cmidrule(lr){2-2}\cmidrule(lr){3-6}\cmidrule(lr){7-9}
    Method
      & CLIP$\uparrow$
      & RotErr$\downarrow$ & TransErr$\downarrow$ & CamMC$\downarrow$ & ATE$\downarrow$
      & FVD$\downarrow$ & FVD$_{\mathrm{c}}\downarrow$ & FID$\downarrow$ \\
    \midrule
    \multicolumn{9}{@{}l}{\textit{Wan-2.2 5B Scale}} \\
    ReCamMaster~\cite{bai2025recammaster}     & 24.97  & 0.131 & 0.297 & 0.402 & 0.322 & 874.30 & 890.52 & 62.53 \\
    ReRoPE~\cite{li2026rerope}                & 25.20  & 0.137 & 0.241 & 0.366 & 0.307 & 684.57 & 650.31 & 60.77 \\
    UCPE~\cite{zhang2026ucpe}                 & 25.39  & 0.113 & 0.180 & 0.274 & 0.203 & 703.41 & 755.83 & 61.50 \\
    \method (ours)                            & \textbf{26.05}  & \textbf{0.085} & \textbf{0.174} & \textbf{0.251} & \textbf{0.188} & \textbf{543.17} & \textbf{588.62} & \textbf{57.83} \\
    \midrule
    \multicolumn{9}{@{}l}{\textit{Wan-2.2 14B Scale}} \\
    ReCamMaster~\cite{bai2025recammaster}     & 25.31      & 0.109 & 0.221 & 0.335 & 0.286 & 675.23  & 697.10 & 59.21     \\
    ReRoPE~\cite{li2026rerope}                & 25.85  & 0.114 & 0.185 & 0.290 & 0.243 & 493.30 & 525.61 & 49.18 \\
    UCPE~\cite{zhang2026ucpe}                 & 25.72  & 0.082 & 0.163 & 0.238 & 0.180 & 529.42 & 558.70 & 54.75 \\
    \method (ours)                            & \textbf{26.30}  & \textbf{0.058} & \textbf{0.126} & \textbf{0.165} & \textbf{0.144} & \textbf{280.17} & \textbf{354.52} & \textbf{41.01} \\
    \bottomrule
  \end{tabular}
\end{table*}

\subsection{Bilinearity of the Pl\"ucker Reciprocal Product}
\label{sec:app-bilinear}

We restate the elementary fact used in
Section~\ref{sec:bg-plucker}. Let $\rvec_i=(\dvec_i,\mvec_i)$ and
$\rvec_j=(\dvec_j,\mvec_j)$ be two 6-vectors. Define
\begin{equation}
  \plinner{\rvec_i}{\rvec_j} \;=\; \dvec_i\!\cdot\!\mvec_j + \dvec_j\!\cdot\!\mvec_i.
\end{equation}
For any $\alpha,\beta\!\in\!\mathbb{R}$ and any $\rvec'_i$,
\begin{align*}
\plinner{\alpha\rvec_i+\beta\rvec'_i}{\rvec_j}
&= (\alpha\dvec_i+\beta\dvec'_i)\!\cdot\!\mvec_j
 + (\alpha\mvec_i+\beta\mvec'_i)\!\cdot\!\dvec_j \\
&= \alpha\bigl(\dvec_i\!\cdot\!\mvec_j
      +\mvec_i\!\cdot\!\dvec_j\bigr) \\
&\quad
 +\beta\bigl(\dvec'_i\!\cdot\!\mvec_j
      +\mvec'_i\!\cdot\!\dvec_j\bigr) \\
&= \alpha\,\plinner{\rvec_i}{\rvec_j}
 + \beta\,\plinner{\rvec'_i}{\rvec_j}.
\end{align*}
The same holds in the second argument by symmetry. Hence
$\plinner{\cdot}{\cdot}$ is bilinear.

\subsection{Coplanarity Criterion}
\label{sec:app-coplanar}

Two rays $\rvec_i,\rvec_j$ are coplanar (i.e.\ they intersect, are
parallel, or coincide) if and only if
$\plinner{\rvec_i}{\rvec_j}=0$. The proof is standard~\cite{hartley2004multi,pottmann2001computational}
and rests on the observation that
$\dvec_i\!\cdot\!\mvec_j+\dvec_j\!\cdot\!\mvec_i$ is, up to a sign,
the determinant of a $4\!\times\!4$ matrix whose columns are the
homogeneous coordinates of two points on each line; the determinant
vanishes iff the four points are coplanar.

\subsection{SE(3) Invariance}
\label{sec:app-se3}

For a rigid motion $g=(R,\mathbf{t})\!\in\!\mathrm{SE}(3)$, the
action on a ray is
\begin{equation}
  g\!\cdot\!(\dvec,\mvec) \;=\; (R\dvec,\,R\mvec + \mathbf{t}\times R\dvec).
\end{equation}
Substituting into Eq.~\eqref{eq:plucker-inner},
\begin{align*}
\plinner{g\!\cdot\!\rvec_i}{g\!\cdot\!\rvec_j}
&= R\dvec_i\!\cdot\!(R\mvec_j+\mathbf{t}\times R\dvec_j) \\
&\quad
 + R\dvec_j\!\cdot\!(R\mvec_i+\mathbf{t}\times R\dvec_i) \\
&= \dvec_i\!\cdot\!\mvec_j
 + \dvec_j\!\cdot\!\mvec_i \\
&\quad
 + R\dvec_i\!\cdot\!(\mathbf{t}\times R\dvec_j) \\
&\quad
 + R\dvec_j\!\cdot\!(\mathbf{t}\times R\dvec_i).
\end{align*}
The last two terms cancel because the scalar triple product
$a\!\cdot\!(b\times c)$ is antisymmetric in any two of its arguments,
giving
$R\dvec_i\!\cdot\!(\mathbf{t}\times R\dvec_j) = -R\dvec_j\!\cdot\!(\mathbf{t}\times R\dvec_i)$.
Hence
$\plinner{g\!\cdot\!\rvec_i}{g\!\cdot\!\rvec_j}=\plinner{\rvec_i}{\rvec_j}$.

\subsection{Reduction of Term~(D) to the Pl\"ucker Reciprocal Product}
\label{sec:app-flip}

Recall (Eq.~\eqref{eq:term-D}) that, with the Q/K flip,
\begin{equation}
  \mathrm{(D)} = \rvec_i^\top \,M\, \tilde\rvec_j,\qquad
  \rvec_i = (\dvec_i,\mvec_i),\quad
  \tilde\rvec_j = (\mvec_j,\dvec_j),
\end{equation}
where $M=E_q^\top E_k\in\mathbb{R}^{6\times 6}$. Writing
$M=\begin{pmatrix}A & B\\C & D\end{pmatrix}$ with
$A,B,C,D\in\mathbb{R}^{3\times 3}$,
\begin{equation}
  \mathrm{(D)} =
  \dvec_i^\top A\mvec_j + \dvec_i^\top B\dvec_j +
  \mvec_i^\top C\mvec_j + \mvec_i^\top D\dvec_j.
\end{equation}
Setting $A=D=I_3$, $B=C=\mathbf{0}_3$ collapses this to
$\dvec_i\!\cdot\!\mvec_j+\mvec_i\!\cdot\!\dvec_j=\plinner{\rvec_i}{\rvec_j}$.
The choice $E_q=E_k=I_6$ in Proposition~\ref{prop:flip} produces
exactly $M=I_6$, so the full $M$-block decomposition above shows
\emph{which} sub-blocks of $M$ correspond to the reciprocal-product
component and which sub-blocks generalize beyond it: the
off-diagonal blocks $A,D$ control the reciprocal-product-like part,
while the diagonal blocks $B,C$ control direction--direction and
moment--moment couplings that are not present in the classical
reciprocal product.

\subsection{Flipped vs. Unflipped Parameterization Are Linearly Equivalent}
\label{sec:app-flip-reparam}

Let $\rvec=(\dvec,\mvec)\in\mathbb{R}^6$ and let
$P=\bigl(\begin{smallmatrix}0 & I_3 \\ I_3 & 0\end{smallmatrix}\bigr)$
be the orthogonal involution that swaps the $(\dvec,\mvec)$ blocks
($P^2=I_6$, $P^\top=P$). With learnable
$E_q,E_k\in\mathbb{R}^{d\times 6}$, the flipped parameterization in
Eqs.~\eqref{eq:flip-q}--\eqref{eq:flip-k} feeds $\rvec_i$ to $E_q$
and $P\rvec_j$ to $E_k$, while the unflipped variant feeds $\rvec_j$
to $E_k$ directly. The corresponding term~(D) in
Eq.~\eqref{eq:term-D} reads
\[
  \mathrm{(D)}_{\mathrm{flip}}
    = \rvec_i^\top E_q^\top E_k\,P\,\rvec_j,
  \qquad
  \mathrm{(D)}_{\mathrm{no\text{-}flip}}
    = \rvec_i^\top E_q^\top E_k\,\rvec_j.
\]
For unconstrained $E_k$ the map $E_k\mapsto E_k P$ is a bijection on
$\mathbb{R}^{d\times 6}$, so the two parameterizations realize the
same set of bilinear forms in $(\rvec_i,\rvec_j)$. Under iid Gaussian
initialization the columns of $E_k$ and of $E_kP$ also share the
same distribution, so $\ell_2$ weight decay is invariant under the
substitution. The flip is therefore a choice of basis: it singles out
a distinguished symmetric configuration ($E_q=E_k=I_6$) under which
the bilinear form coincides with the Pl\"ucker reciprocal product
(Proposition~\ref{prop:flip}), but it does not change the function
class realizable by the model.

\subsection{Implementation Notes}
\label{sec:app-impl}

\paragraph{Numerical safety.}
We clamp $\|\mvec\|$ at $\epsilon=10^{-6}$ before normalization
(Eq.~\eqref{eq:decompose}) to avoid division by zero in
near-static-camera tokens.

\paragraph{Old-checkpoint compatibility.}
A previous version of the encoding (without PE RMSNorm) shipped
$\alpha_q,\alpha_k$ as zero-dimensional scalars, which are not
compatible with FSDP's flat-parameter constraint that all sharded
tensors have $\geq 1$ dimension. Current checkpoints store these as
shape-$(1,)$ tensors and the loader silently promotes old scalars at
load time.

\paragraph{Two-layer MLP form of $E_q,E_k$.}
When the optional MLP form
$E_q\!:\!\mathbb{R}^7\!\to\!\mathbb{R}^h\!\to\!\mathbb{R}^d$ is used,
zero-initializing both layers would produce a permanently dead
gradient (the GELU after a zero linear is identically zero, and so
is its Jacobian). We therefore zero only the output layer and apply
Kaiming initialization to the input layer, which keeps the geometric
branch exactly zero at step $0$ while leaving its gradient non-zero.

\subsection{Reference Results under Rescaled-Alignment}
\label{sec:app-rescaled}
We briefly recap why the main paper reports pose metrics on the
\emph{raw, unscaled} ViPE trajectories. A camera-conditioned video
model is meant to follow a user-specified trajectory
\emph{including its absolute scale}; once the predicted and the
ground-truth trajectories are rescaled to a common norm at
evaluation time, the resulting numbers no longer measure whether
the model honored that scale, only whether the relative shape of
the trajectory was reproduced. Two practical consequences follow.
First, a degenerate solution that produces only a tiny global
motion can still score well under the conventional protocol,
because once both trajectories are stretched to the same norm the
remaining shape error is small even though the absolute motion is
essentially absent; the raw protocol exposes this failure mode
directly. Second, rescaled pose numbers are implicitly expressed in
units of each clip's own ground-truth norm and are therefore
difficult to compare across datasets with different scale
conventions (SfM-normalized, SLAM-internal, metric), whereas the
raw protocol keeps the unit fixed to ViPE's metric output. We
therefore adopt the raw protocol throughout the main paper, and
report rescaled-alignment numbers here only as a cross-reference to
prior camera-controllability work.
Concretely, Table~\ref{tab:main-rescaled} repeats the comparison
from Section~\ref{sec:exp-main} after normalizing the predicted and
the ground-truth ViPE trajectories to a common translation norm
before computing pose metrics. Everything else --- held-out split,
trajectory estimator, sampler configuration, and the
video-quality metrics CLIP, FVD, FVD$_{\mathrm{c}}$, and FID --- is
identical to Table~\ref{tab:main}; only the alignment step differs.
Because rescaling absorbs absolute-scale errors, the four pose
metrics are systematically smaller than their unscaled counterparts
in the main table. \method remains best on every metric in this
setting as well, confirming that its advantage under the unscaled
protocol is not an artifact of the alignment choice.

\section{Additional Evaluation}
\label{sec:app-additional-eval}

\subsection{Dynamic-Scene Video Quality}
\label{sec:app-vbench}

\paragraph{Purpose and setting.}
The camera-control metrics in the main paper do not directly test
whether a model preserves the backbone's ability to generate dynamic
content. A method could follow the camera trajectory while producing
overly static subjects, or introduce motion at the cost of temporal
coherence and visual quality. We therefore conduct an additional
dynamic-scene evaluation on a comprehensive held-out set drawn from
PanShot, OmniWorld, and Open-Sora-Plan. The evaluation videos do not
overlap with the training set. All methods use the Wan2.2-14B
backbone and receive the same conditioning inputs and target camera
trajectories. We evaluate the generated videos with the official
VBench~\cite{huang2024vbench} implementation under its default configuration and apply the
same metrics to the corresponding real videos, reported as ground
truth in Table~\ref{tab:app-vbench}.

We report six complementary VBench dimensions. Subject and Background
Consistency measure temporal preservation of the foreground and
scene, Motion Smoothness measures temporal regularity, and Dynamic
Degree directly measures the amount of generated motion. Aesthetic
Quality and Imaging Quality measure perceptual quality. Higher is
better for every metric.

\begin{table*}[t]
  \centering
  \caption{Dynamic-scene evaluation on the held-out PanShot,
    OmniWorld, and Open-Sora-Plan~\cite{lin2024open} mixture using Wan2.2-14B and the
    official VBench default configuration. ``GT'' denotes the real
    evaluation videos. All generated methods receive identical
    conditioning inputs and target camera trajectories. Higher is
    better for all metrics; bold marks the best generated result.}
  \label{tab:app-vbench}
  \small
  \resizebox{\textwidth}{!}{%
  \begin{tabular}{lcccccc}
    \toprule
    Method
      & Subject Consistency $\uparrow$
      & Background Consistency $\uparrow$
      & Motion Smoothness $\uparrow$
      & Dynamic Degree $\uparrow$
      & Aesthetic Quality $\uparrow$
      & Imaging Quality $\uparrow$ \\
    \midrule
    GT                                        & 0.935 & 0.942 & 0.984 & 0.863 & 0.558 & 0.692 \\
    \midrule
    ReCamMaster~\cite{bai2025recammaster}    & 0.782 & 0.841 & 0.909 & 0.490 & 0.386 & 0.451 \\
    PRoPE~\cite{li2025prope}                 & 0.855 & 0.903 & 0.955 & 0.693 & 0.451 & 0.606 \\
    UCPE~\cite{zhang2026ucpe}                & 0.897 & 0.928 & 0.979 & 0.752 & 0.508 & 0.622 \\
    \method (ours)                           & \textbf{0.918} & \textbf{0.945} & \textbf{0.983} & \textbf{0.855} & \textbf{0.561} & \textbf{0.689} \\
    \bottomrule
  \end{tabular}}
\end{table*}

\begin{table*}[t]
  \centering
  \caption{Generalization beyond RealEstate10K on a held-out mixture
    of PanShot, OmniWorld, and Open-Sora-Plan using Wan2.2-5B. Pose
    cells report raw/rescaled-alignment results; FVD metrics are
    unaffected by trajectory alignment. Both methods use identical
    conditioning inputs, target trajectories, and evaluation
    settings. Lower is better for all metrics.}
  \label{tab:app-beyond-re10k}
  \small
  \begin{tabular}{lcccccc}
    \toprule
    Method
      & RotErr $\downarrow$
      & TransErr $\downarrow$
      & CamMC $\downarrow$
      & ATE $\downarrow$
      & FVD $\downarrow$
      & FVD$_{\mathrm{c}}$ $\downarrow$ \\
    \midrule
    UCPE~\cite{zhang2026ucpe}
      & 0.136/0.136 & 1.590/0.338 & 1.704/0.393 & 1.881/0.379 & 694.5 & 722.1 \\
    \method (ours)
      & \textbf{0.089/0.089} & \textbf{1.164/0.220}
      & \textbf{1.091/0.279} & \textbf{1.217/0.208}
      & \textbf{529.0} & \textbf{550.8} \\
    \bottomrule
  \end{tabular}
\end{table*}

\paragraph{Results.}
As shown in Table~\ref{tab:app-vbench}, \method ranks first among the
generated methods on all six dimensions. Most notably, its Dynamic
Degree reaches $0.855$, close to the real-video value of $0.863$ and
substantially above the strongest baseline value of $0.752$. At the
same time, \method nearly matches the real videos in Motion
Smoothness ($0.983$ versus $0.984$), Aesthetic Quality ($0.561$
versus $0.558$), and Imaging Quality ($0.689$ versus $0.692$).
Subject and Background Consistency also improve over all baselines.
These results show that the gains in camera controllability do not
come from suppressing scene motion: \method preserves dynamic content
while maintaining temporal coherence and perceptual quality.

\subsection{Generalization beyond RealEstate10K}
\label{sec:app-beyond-re10k}

\paragraph{Purpose and setting.}
The main evaluation uses the standard RealEstate10K benchmark, whose
videos are dominated by real-estate walkthroughs. To test whether the
advantage of \method transfers to broader scene and motion
distributions, we evaluate on the same held-out PanShot, OmniWorld,
and Open-Sora-Plan mixture used in
Section~\ref{sec:app-vbench}. This set has no overlap with the
training videos and includes more diverse visual content, dynamic
scenes, and camera trajectories than RealEstate10K. Both UCPE and
\method use the same Wan2.2-5B backbone and are evaluated from the
same conditioning frames and target trajectories with the sampling
and pose-estimation protocol of Section~\ref{sec:exp-setup}.

Following the main evaluation, we estimate the generated and
ground-truth trajectories with ViPE and report RotErr, TransErr,
CamMC, and ATE. Each pose cell in
Table~\ref{tab:app-beyond-re10k} gives the raw result followed by the
rescaled-alignment result. We additionally report full-frame FVD and
center-cropped FVD$_{\mathrm{c}}$ to measure distribution-level video
quality.

\paragraph{Results.}
Table~\ref{tab:app-beyond-re10k} shows that \method remains better
than UCPE on every camera-control and distribution metric. Under the
raw protocol, \method reduces RotErr, TransErr, CamMC, and ATE by
$35\%$, $27\%$, $36\%$, and $35\%$, respectively. The advantage is
preserved after conventional trajectory rescaling, showing that it
does not arise only from absolute motion scale. \method also lowers
FVD from $694.5$ to $529.0$ and FVD$_{\mathrm{c}}$ from $722.1$ to
$550.8$, both by approximately $24\%$. The consistent improvement on
this broader held-out distribution indicates that the ray-coordinate
encoding generalizes beyond the real-estate scenes used by the main
benchmark.

\begin{table}[!htbp]
  \centering
  \caption{Inference efficiency on Wan2.2-TI2V-5B. Step latency is
    averaged over the 50 denoising steps of a complete inference run.
    All methods use the same input and inference configuration.
    Lower is better for both metrics.}
  \label{tab:app-efficiency}
  \small
  \setlength{\tabcolsep}{4pt}
  \resizebox{\linewidth}{!}{%
  \begin{tabular}{lccc}
    \toprule
    Method & Camera integration
      & Step latency (s) $\downarrow$
      & Peak memory (GB) $\downarrow$ \\
    \midrule
    UCPE~\cite{zhang2026ucpe}
      & Compressed parallel branch & 27.6 & 21.8 \\
    PRoPE~\cite{li2025prope}
      & Full parallel branch & 34.6 & 22.3 \\
    \method (ours)
      & Additive Q/K injection & \textbf{24.4} & \textbf{11.6} \\
    \bottomrule
  \end{tabular}}
\end{table}

\subsection{Runtime and Memory}
\label{sec:app-efficiency}

\paragraph{Purpose and setting.}
We compare the inference cost introduced by different camera-
conditioning mechanisms on the Wan2.2-TI2V-5B backbone. All methods
generate videos at \(480\times832\) resolution with 81 frames under
the same inference configuration. Step latency is the mean wall-clock
time of one denoising step, averaged over the 50 denoising steps of a
complete inference run. Peak memory is measured during the same run.

\paragraph{Results.}
\method projects the ray features directly into the existing queries
and keys, requiring neither a parallel branch nor an additional
attention map. Its added per-token computation is
\(\mathcal{O}(Nd)\), which is small relative to the
\(\mathcal{O}(N^2d)\) self-attention computation. UCPE instead uses a
parallel geometry branch with an attention-compression ratio of
eight. For PRoPE, direct substitution of the pretrained positional
transformation did not yield stable fine-tuning under the matched
budget; we therefore use a zero-initialized full parallel branch for
the trainable video-DiT configuration reported here.

As shown in Table~\ref{tab:app-efficiency}, \method has the lowest
inference cost. It reduces step latency by \(12\%\) relative to UCPE
and \(30\%\) relative to PRoPE. Its peak memory is \(11.6\) GB,
compared with \(21.8\) GB for UCPE and \(22.3\) GB for PRoPE,
reducing memory consumption by approximately \(47\%\) and \(48\%\),
respectively. These results show that the direct Q/K injection
retains the efficiency of the pretrained attention path while
avoiding the cost of a parallel camera-conditioning branch.

%% file: sections/05_training.tex
\section{Cross-Domain Training Strategy}
\label{sec:training}

This section describes the data sources we use, the steps we take to
make their camera-translation scales comparable, and the optimization
recipe we apply on top of the pretrained backbone.

\subsection{Data Sources}
\label{sec:training-data}

We train on a concatenation of four datasets that span quite different
camera-motion distributions and pose-estimation pipelines:

\begin{itemize}
  \item \textbf{RealEstate10K} (RE10K)~\cite{zhou2018re10k}: about 10K
    real-estate walkthrough clips with COLMAP-style SfM
    poses~\cite{schoenberger2016sfm}. Smooth, primarily forward
    camera motion through indoor scenes; the canonical
    benchmark for camera-controllable video synthesis.
  \item \textbf{DL3DV}~\cite{ling2024dl3dv}: a more diverse
    multi-scene dataset with both indoor and outdoor scenes and a
    wider range of camera motions; poses provided by the dataset.
  \item \textbf{PanShot}~\cite{zhang2026ucpe} (in-house): an internal dataset of synthesized
    panoramic shots, where camera trajectories are known by
    construction (we use them as ground-truth) and the field of view
    is known per clip.
  \item \textbf{OmniWorld}~\cite{zhou2025omniworld}: open-world video with
    relatively dynamic content, with camera extrinsics estimated by
    DROID-SLAM~\cite{teed2021droidslam}. We use the released text
    captions as conditioning prompts. DROID-SLAM applies an internal
    scale normalization that is unrelated to either metric scale or
    the SfM normalization of RE10K/DL3DV.
\end{itemize}

We do not perform any further re-estimation of camera pose; \method
treats the per-frame extrinsics and intrinsics as inputs and is
agnostic to which estimator produced them. The four data sources span
SfM poses, deep SLAM poses, and ground-truth-by-construction poses,
which provides a stress test for whether the encoding generalizes
across pose distributions.

\subsection{Per-Dataset Trajectory-Scale Alignment}
\label{sec:training-trajscale}

The four data sources differ substantially in absolute translation
magnitude. To make their Pl\"ucker moments comparable in distribution,
we apply a single per-dataset multiplicative factor
$\eta\!\in\!\mathbb{R}_{>0}$ to the camera positions:
\begin{equation}
  \tilde\ovec_t \;=\; \eta\cdot \ovec_t,
  \qquad
  \tilde\mvec_i = \tilde\ovec_t\!\times\!\dvec_i = \eta\cdot\mvec_i.
\end{equation}
The values of $\eta$ for each dataset are chosen so that the median
moment magnitude $\mathrm{median}(\|\mvec\|)$ over a sample of clips
falls in a comparable range across datasets; in our default
configuration $\eta_{\mathrm{RE10K}}=\eta_{\mathrm{DL3DV}}=\eta_{\mathrm{PanShot}}=1.0$
and $\eta_{\mathrm{OmniWorld}}=20.0$, the latter chosen to bring
DROID-SLAM's internally normalized translations into approximately the
same band as RE10K. Concrete moment-magnitude statistics are reported
in Section~\ref{sec:experiments}.

\subsection{Optional Per-Clip Near-Depth Normalization}
\label{sec:training-neardepth}

A constant per-dataset $\eta$ does not absorb \emph{within}-dataset
variation: a long-corridor RE10K clip and a tight-room RE10K clip can
still differ by an order of magnitude in scene depth. When a per-clip
near-depth estimate $z_n$ is available, we additionally divide the
clip's translation by $z_n$:
\begin{equation}
  \tilde\ovec_t \;=\; \frac{\eta}{z_n}\cdot \ovec_t.
\end{equation}
The near-depth estimate $z_n$ is obtained offline (from a depth
estimator on a small set of frames) and serves as a per-clip proxy
for the scene scale. When $z_n$ is unavailable for a clip we fall
back to $\eta$ alone. For OmniWorld we use only $\eta$ in this work.

These two normalizations (per-dataset $\eta$ and per-clip $1/z_n$)
together aim to make the moment-magnitude distribution roughly
homogeneous across the training mixture. The \ngi mechanism
(Section~\ref{sec:method-ngi}) provides robustness for the residual
within-distribution variation that this preprocessing does not remove.

\subsection{Optimization Recipe}
\label{sec:training-opt}

\paragraph{Backbone and conditioning.}
We start from a pretrained 5B-parameter Wan2.2-TI2V-5B
backbone~\cite{wan2025wan} and use it in its image-to-video mode (the
first frame of the training clip is conditioned via the standard
first-frame VAE-fuse path of the backbone). All experiments are
trained at $480\!\times\!832$ spatial resolution and $T\!=\!81$ raw
frames (corresponding to $T_z\!=\!21$ latent frames).

\paragraph{Zero initialization.}
The new Q/K geometry branch is zero-initialized, so
$\mathrm{pe}^q\!=\!\mathrm{pe}^k\!=\!0$ at step~0 and the network
matches the pretrained DiT. PE RMSNorm weights start at $1$,
scale-gate biases at $0$ (so $g\!\approx\!0.5$), and $\alpha$ starts at
$0$. When $E_q,E_k$ are implemented as two-layer MLPs, we zero only the
output layer and use Kaiming initialization for the input layer.

\paragraph{Trainable parameters and optimizer.}
We jointly train the new \method modules ($E_q,E_k$, PE RMSNorms,
scale gates, $\alpha$) together with the backbone self-attention,
FFN, and modulation-norm layers; text cross-attention, the VAE, and
the text encoder remain frozen. The newly added \method parameters account for
less than $0.1\%$ of the model. We use AdamW with a single learning
rate of $2\!\times\!10^{-5}$ across all trainable groups,
$(\beta_1,\beta_2)\!=\!(0.9,0.999)$, $\epsilon\!=\!10^{-8}$, weight
decay $10^{-2}$, and a cosine schedule that anneals to $10\%$ of the
starting value at $T_{\max}$ steps. We tried per-group asymmetric
learning rates that protect the pretrained weights more aggressively
but observed no consistent improvement, so we kept the simpler
single-LR recipe.

\paragraph{Mixed precision and parallelism.}
We train in BF16 with FP32 master weights, gradient checkpointing on
the DiT blocks, and DDP across GPUs. Gradient clipping by norm at
$1.0$ is applied at the DDP root level.

\paragraph{Validation-time generation.}
Every $5{,}000$ steps we run inference on a small held-out validation
set (a 100-clip RE10K-test split) and save (i) the per-step training
and validation loss, (ii) the synthesized videos, and (iii) ground-truth
videos for visual comparison. Validation videos are generated with the
standard 50-step CFG sampler at $\mathrm{cfg\_scale}=5.0$.

%% file: references.bib
@inproceedings{ho2022vdm,
  title={Video Diffusion Models},
  author={Ho, Jonathan and Salimans, Tim and Gritsenko, Alexey and Chan, William and Norouzi, Mohammad and Fleet, David J.},
  booktitle={NeurIPS},
  year={2022}
}

@article{blattmann2023svd,
  title={Stable video diffusion: Scaling latent video diffusion models to large datasets},
  author={Blattmann, Andreas and Dockhorn, Tim and Kulal, Sumith and Mendelevitch, Daniel and Kilian, Maciej and Lorenz, Dominik and Levi, Yam and English, Zion and Voleti, Vikram and Letts, Adam and Jampani, Varun and Rombach, Robin},
  journal={arXiv:2311.15127},
  year={2023}
}

@inproceedings{yang2025cogvideox,
  title={Cogvideox: Text-to-video diffusion models with an expert transformer},
  author={Yang, Zhuoyi and Teng, Jiayan and Zheng, Wendi and Ding, Ming and Huang, Shiyu and Xu, Jiazheng and Yang, Yuanming and Hong, Wenyi and Zhang, Xiaohan and Feng, Guanyu and Yin, Da and Gu, Xiaotao and Zhang, Yuxuan and Wang, Weihan and Cheng, Yean and Liu, Ting and Xu, Bin and Dong, Yuxiao and Tang, Jie},
  booktitle={ICLR},
  year={2025}
}

@article{kong2024hunyuanvideo,
  title={{HunyuanVideo}: A Systematic Framework for Large Video Generative Models},
  author={Kong, Weijie and Tian, Qi and Zhang, Zijian and Min, Rox and Dai, Zuozhuo and Zhou, Jin and Xiong, Jiangfeng and Li, Xin and Wu, Bo and others},
  journal={arXiv:2412.03603},
  year={2024}
}

@article{polyak2024moviegen,
  title={Movie {Gen}: A Cast of Media Foundation Models},
  author={Polyak, Adam and Zohar, Amit and Brown, Andrew and Tjandra, Andros and Sinha, Animesh and others},
  journal={arXiv:2410.13720},
  year={2024}
}

@article{wan2025wan,
  title={Wan: Open and advanced large-scale video generative models},
  author={Wan, Team and Wang, Ang and Ai, Baole and Wen, Bin and Mao, Chaojie and Xie, Chen-Wei and Chen, Di and Yu, Feiwu and Zhao, Haiming and Yang, Jianxiao and others},
  journal={arXiv:2503.20314},
  year={2025}
}

@article{su2021rope,
  title={{RoFormer}: Enhanced Transformer with Rotary Position Embedding},
  author={Su, Jianlin and Ahmed, Murtadha and Lu, Yu and Pan, Shengfeng and Wen, Bo and Liu, Yunfeng},
  journal={Neurocomputing},
  year={2024}
}

@inproceedings{heo2024rotary,
  title={Rotary Position Embedding for Vision Transformer},
  author={Heo, Byeongho and Park, Song and Han, Dongyoon and Yun, Sangdoo},
  booktitle={ECCV},
  year={2024}
}

@inproceedings{henry2020qknorm,
  title={Query-Key Normalization for Transformers},
  author={Henry, Alex and Dachapally, Prudhvi Raj and Pawar, Shubham Shantaram and Chen, Yuxuan},
  booktitle={Findings of EMNLP},
  year={2020}
}

@inproceedings{li2025prope,
  title={Cameras as Relative Positional Encoding},
  author={Li, Ruilong and Yi, Brent and Liu, Junchen and Gao, Hang and Ma, Yi and Kanazawa, Angjoo},
  booktitle={NeurIPS},
  year={2025}
}

@inproceedings{he2025cameractrl,
  title={Cameractrl: Enabling camera control for video diffusion models},
  author={He, Hao and Xu, Yinghao and Guo, Yuwei and Wetzstein, Gordon and Dai, Bo and Li, Hongsheng and Yang, Ceyuan},
  booktitle={ICLR},
  year={2025}
}

@inproceedings{wang2024motionctrl,
  title={{MotionCtrl}: A Unified and Flexible Motion Controller for Video Generation},
  author={Wang, Zhouxia and Yuan, Ziyang and Wang, Xintao and Li, Yaowei and Chen, Tianshui and Xia, Menghan and Luo, Ping and Shan, Ying},
  booktitle={SIGGRAPH},
  year={2024}
}

@article{xu2024camco,
  title={{CamCo}: Camera-Controllable {3D}-Consistent Image-to-Video Generation},
  author={Xu, Dejia and Nie, Weili and Liu, Chao and Liu, Sifei and Kautz, Jan and Wang, Zhangyang and Vahdat, Arash},
  journal={arXiv:2406.02509},
  year={2024}
}

@inproceedings{bahmani2025ac3d,
  title={{AC3D}: Analyzing and Improving 3D Camera Control in Video Diffusion Transformers},
  author={Bahmani, Sherwin and Skorokhodov, Ivan and Qian, Guocheng and Siarohin, Aliaksandr and Menapace, Willi and Tagliasacchi, Andrea and Lindell, David B. and Tulyakov, Sergey},
  booktitle={CVPR},
  year={2025}
}

@inproceedings{bahmani2025vd3d,
  title={{VD3D}: Taming Large Video Diffusion Transformers for 3D Camera Control},
  author={Bahmani, Sherwin and Skorokhodov, Ivan and Siarohin, Aliaksandr and Menapace, Willi and Qian, Guocheng and Vasilkovsky, Michael and Lee, Hsin-Ying and Wang, Chaoyang and Zou, Jiaxu and Tagliasacchi, Andrea and Lindell, David B. and Tulyakov, Sergey},
  booktitle={ICLR},
  year={2025}
}

@article{yu2024viewcrafter,
  title={{ViewCrafter}: Taming Video Diffusion Models for High-Fidelity Novel View Synthesis},
  author={Yu, Wangbo and Xing, Jinbo and Yuan, Li and Hu, Wenbo and Li, Xiaoyu and Huang, Zhipeng and Gao, Xiangjun and Wong, Tien-Tsin and Shan, Ying and Tian, Yonghong},
  journal={TPAMI},
  year={2025}
}

@inproceedings{yu2025trajectorycrafter,
  title={{TrajectoryCrafter}: Redirecting Camera Trajectory for Monocular Videos via Diffusion Models},
  author={Yu, Wangbo and Hu, Wenbo and Xing, Jinbo and Shan, Ying},
  booktitle={ICCV},
  year={2025}
}

@inproceedings{bai2025recammaster,
  title={{ReCamMaster}: Camera-Controlled Generative Rendering from a Single Video},
  author={Bai, Jianhong and Xia, Menghan and Fu, Xiao and Wang, Xintao and Mu, Lianrui and Cao, Jinwen and Liu, Zuozhu and Hu, Haoji and Bai, Xiang and Wan, Pengfei and Zhang, Di},
  booktitle={ICCV},
  year={2025}
}

@inproceedings{bian2025gsdit,
  title={{GS-DiT}: Advancing Video Generation with Dynamic {3D} Gaussian Fields through Efficient Dense {3D} Point Tracking},
  author={Bian, Weikang and Huang, Zhaoyang and Shi, Xiaoyu and Li, Yijin and Wang, Fu-Yun and Li, Hongsheng},
  booktitle={CVPR},
  year={2025}
}

@inproceedings{zhang2026ucpe,
  title={Unified Camera Positional Encoding for Controlled Video Generation},
  author={Zhang, Cheng and Li, Boying and Wei, Meng and Cao, Yan-Pei and Cruz Gambardella, Camilo and Phung, Dinh and Cai, Jianfei},
  booktitle={CVPR},
  year={2026}
}

@article{li2026rerope,
  title={{ReRoPE}: Repurposing {RoPE} for Relative Camera Control},
  author={Li, Chunyang and Yang, Yuanbo and Shao, Jiahao and Zhou, Hongyu and Schwarz, Katja and Liao, Yiyi},
  journal={arXiv:2602.08068},
  year={2026}
}

@inproceedings{gortler1996lumigraph,
  title={The {Lumigraph}},
  author={Gortler, Steven J. and Grzeszczuk, Radek and Szeliski, Richard and Cohen, Michael F.},
  booktitle={SIGGRAPH},
  year={1996}
}

@inproceedings{attal2022learning,
  title={Learning Neural Light Fields with Ray-Space Embedding Networks},
  author={Attal, Benjamin and Huang, Jia-Bin and Zollh{\"o}fer, Michael and Kopf, Johannes and Kim, Changil},
  booktitle={CVPR},
  year={2022}
}

@inproceedings{suhail2022light,
  title={Light Field Neural Rendering},
  author={Suhail, Mohammed and Esteves, Carlos and Sigal, Leonid and Makadia, Ameesh},
  booktitle={CVPR},
  year={2022}
}

@inproceedings{zhang2024camerasasrays,
  title={Cameras as Rays: Pose Estimation via Ray Diffusion},
  author={Zhang, Jason Y. and Lin, Amy and Kumar, Moneish and Yang, Tzu-Hsuan and Ramanan, Deva and Tulsiani, Shubham},
  booktitle={ICLR},
  year={2024}
}

@book{hartley2004multi,
  title={Multiple View Geometry in Computer Vision},
  author={Hartley, Richard and Zisserman, Andrew},
  publisher={Cambridge University Press},
  year={2004}
}

@book{pottmann2001computational,
  title={Computational Line Geometry},
  author={Pottmann, Helmut and Wallner, Johannes},
  publisher={Springer},
  year={2001}
}

@inproceedings{schoenberger2016sfm,
  title={Structure-from-Motion Revisited},
  author={Sch{\"o}nberger, Johannes Lutz and Frahm, Jan-Michael},
  booktitle={CVPR},
  year={2016}
}

@inproceedings{teed2021droidslam,
  title={{DROID-SLAM}: Deep Visual {SLAM} for Monocular, Stereo, and {RGB-D} Cameras},
  author={Teed, Zachary and Deng, Jia},
  booktitle={NeurIPS},
  year={2021}
}

@article{zhou2018re10k,
  title={Stereo Magnification: Learning View Synthesis using Multiplane Images},
  author={Zhou, Tinghui and Tucker, Richard and Flynn, John and Fyffe, Graham and Snavely, Noah},
  journal={ACM TOG},
  year={2018}
}

@inproceedings{ling2024dl3dv,
  title={Dl3dv-10k: A large-scale scene dataset for deep learning-based 3d vision},
  author={Ling, Lu and Sheng, Yichen and Tu, Zhi and Zhao, Wentian and Xin, Cheng and Wan, Kun and Yu, Lantao and Guo, Qianyu and Yu, Zixun and Lu, Yawen and Li, Xuanmao and Sun, Xingpeng and Ashok, Rohan and Mukherjee, Aniruddha and Kang, Hao and Kong, Xiangrui and Hua, Gang and Zhang, Tianyi and Benes, Bedrich and Bera, Aniket},
  booktitle={CVPR},
  year={2024}
}

@article{huang2025vipe,
  title={Vipe: Video pose engine for 3d geometric perception},
  author={Huang, Jiahui and Zhou, Qunjie and Rabeti, Hesam and Korovko, Aleksandr and Ling, Huan and Ren, Xuanchi and Shen, Tianchang and Gao, Jun and Slepichev, Dmitry and Lin, Chen-Hsuan and Ren, Jiawei and Xie, Kevin and Biswas, Joydeep and Leal-Taix{\'e}, Laura and Fidler, Sanja},
  journal={arXiv:2508.10934},
  year={2025}
}

@inproceedings{zhou2025omniworld,
  title={{OmniWorld}: A Multi-Domain and Multi-Modal Dataset for 4D World Modeling},
  author={Zhou, Yang and Wang, Yifan and Zhou, Jianjun and Chang, Wenzheng and Guo, Haoyu and Li, Zizun and Ma, Kaijing and Li, Xinyue and Wang, Yating and Zhu, Haoyi and Liu, Mingyu and Liu, Dingning and Yang, Jiange and Fu, Zhoujie and Chen, Junyi and Shen, Chunhua and Pang, Jiangmiao and Zhang, Kaipeng and He, Tong},
  booktitle={ICLR},
  year={2026}
}

@inproceedings{ho2020ddpm,
  title={Denoising Diffusion Probabilistic Models},
  author={Ho, Jonathan and Jain, Ajay and Abbeel, Pieter},
  booktitle={NeurIPS},
  year={2020}
}

@inproceedings{lipman2023flow,
  title={Flow Matching for Generative Modeling},
  author={Lipman, Yaron and Chen, Ricky T.~Q. and Ben-Hamu, Heli and Nickel, Maximilian and Le, Matthew},
  booktitle={ICLR},
  year={2023}
}

@inproceedings{vaswani2017attention,
  title={Attention is All You Need},
  author={Vaswani, Ashish and Shazeer, Noam and Parmar, Niki and Uszkoreit, Jakob and Jones, Llion and Gomez, Aidan N. and Kaiser, {\L}ukasz and Polosukhin, Illia},
  booktitle={NeurIPS},
  year={2017}
}

@inproceedings{unterthiner2019fvd,
  title={{FVD}: A New Metric for Video Generation},
  author={Unterthiner, Thomas and van Steenkiste, Sjoerd and Kurach, Karol and Marinier, Raphael and Michalski, Marcin and Gelly, Sylvain},
  booktitle={ICLR Workshops},
  year={2019}
}

@article{durrant2006simultaneous,
  title={Simultaneous localization and mapping: part I},
  author={Durrant-Whyte, Hugh and Bailey, Tim},
  journal={IEEE Robotics and Automation Magazine},
  year={2006}
}

@inproceedings{gu2025diffusion,
  title={Diffusion as shader: 3d-aware video diffusion for versatile video generation control},
  author={Gu, Zekai and Yan, Rui and Lu, Jiahao and Li, Peng and Dou, Zhiyang and Si, Chenyang and Dong, Zhen and Liu, Qifeng and Lin, Cheng and Liu, Ziwei and Wang, Wenping and Liu, Yuan},
  booktitle={SIGGRAPH},
  year={2025}
}

@inproceedings{zhang2019root,
  title={Root mean square layer normalization},
  author={Zhang, Biao and Sennrich, Rico},
  booktitle={NeurIPS},
  year={2019}
}

@inproceedings{miyato2024gta,
  title={{GTA}: A Geometry-Aware Attention Mechanism for Multi-View Transformers},
  author={Miyato, Takeru and Jaeger, Bernhard and Welling, Max and Geiger, Andreas},
  booktitle={ICLR},
  year={2024}
}

@inproceedings{burgert2025flow,
  title={Go-with-the-Flow: Motion-Controllable Video Diffusion Models Using Real-Time Warped Noise},
  author={Burgert, Ryan and Xu, Yuancheng and Xian, Wenqi and Pilarski, Oliver and Clausen, Pascal and He, Mingming and Ma, Li and Deng, Yitong and Li, Lingxiao and Mousavi, Mohsen and Ryoo, Michael and Debevec, Paul and Yu, Ning},
  booktitle={CVPR},
  year={2025}
}

@inproceedings{zhang2023controlnet,
  title={Adding Conditional Control to Text-to-Image Diffusion Models},
  author={Zhang, Lvmin and Rao, Anyi and Agrawala, Maneesh},
  booktitle={ICCV},
  year={2023}
}

@incollection{adelson1991plenoptic,
  title={The Plenoptic Function and the Elements of Early Vision},
  author={Adelson, Edward H. and Bergen, James R.},
  booktitle={Computational Models of Visual Processing},
  publisher={MIT Press},
  year={1991}
}

@inproceedings{levoy1996light,
  title={Light Field Rendering},
  author={Levoy, Marc and Hanrahan, Pat},
  booktitle={SIGGRAPH},
  year={1996}
}

@inproceedings{sitzmann2021lfns,
  title={Light Field Networks: Neural Scene Representations with Single-Evaluation Rendering},
  author={Sitzmann, Vincent and Rezchikov, Semon and Freeman, William T. and Tenenbaum, Joshua B. and Durand, Fr{\'e}do},
  booktitle={NeurIPS},
  year={2021}
}

@inproceedings{dai2019transformerxl,
  title={Transformer-{XL}: Attentive Language Models beyond a Fixed-Length Context},
  author={Dai, Zihang and Yang, Zhilin and Yang, Yiming and Carbonell, Jaime and Le, Quoc V. and Salakhutdinov, Ruslan},
  booktitle={ACL},
  year={2019}
}

@article{raffel2020t5,
  title={Exploring the Limits of Transfer Learning with a Unified Text-to-Text Transformer},
  author={Raffel, Colin and Shazeer, Noam and Roberts, Adam and Lee, Katherine and Narang, Sharan and Matena, Michael and Zhou, Yanqi and Li, Wei and Liu, Peter J.},
  journal={JMLR},
  year={2020}
}

@article{xu2026ucm,
  title={{UCM}: Unified Modeling of Camera Control and Memory with Time-aware Positional Encoding Warping for World Models},
  author={Xu, Tianxing and Wang, Zixuan and Wang, Guangyuan and Hu, Li and Zhang, Zhongyi and Zhang, Peng and Zhang, Bang and Zhang, Songhai},
  journal={arXiv:2602.22960},
  year={2026}
}

@article{wang2026warphistory,
  title={Warp-as-History: Generalizable Camera-Controlled Video Generation from One Training Video},
  author={Wang, Yifan and He, Tong},
  journal={arXiv:2605.15182},
  year={2026}
}

@inproceedings{bai2025pefield,
  title={Positional Encoding Field},
  author={Bai, Yunpeng and Li, Haoxiang and Huang, Qixing},
  booktitle={ICLR},
  year={2026}
}

@inproceedings{kong2024eschernet,
  title={{EscherNet}: A Generative Model for Scalable View Synthesis},
  author={Kong, Xin and Liu, Shikun and Lyu, Xiaoyang and Taher, Marwan and Qi, Xiaojuan and Davison, Andrew J.},
  booktitle={CVPR},
  year={2024}
}

@article{xiang2026viewrope,
  title={Geometry-Aware Rotary Position Embedding for Consistent Video World Model},
  author={Xiang, Chendong and Liu, Jiajun and Zhang, Jintao and Yang, Xiao and Fang, Zhengwei and Wang, Shizun and Wang, Zijun and Zou, Yingtian and Su, Hang and Zhu, Jun},
  journal={arXiv:2602.07854},
  year={2026}
}

@inproceedings{park2026redirector,
  title={{ReDirector}: Creating Any-Length Video Retakes with Rotary Camera Encoding},
  author={Park, Byeongjun and Kim, Byung-Hoon and Chung, Hyungjin and Ye, Jong Chul},
  booktitle={CVPR},
  year={2026}
}

@article{jin2026crepe,
  title={{CRePE}: Curved Ray Expectation Positional Encoding for Unified-Camera-Controlled Video Generation},
  author={Jin, Seonghyun and Kim, Youngmin and Park, Sunwoo and Ye, Jong Chul},
  journal={arXiv:2605.12938},
  year={2026}
}

@inproceedings{wu2026rayrope,
  title={{RayRoPE}: Projective Ray Positional Encoding for Multi-view Attention},
  author={Wu, Yu and Jeon, Minsik and Chang, Jen-Hao Rick and Tuzel, Oncel and Tulsiani, Shubham},
  booktitle={ECCV},
  year={2026}
}

@inproceedings{xie2026raynova,
  title={{RAYNOVA}: Scale-Temporal Autoregressive World Modeling in Ray Space},
  author={Xie, Yichen and Peng, Chensheng and Abdelfattah, Mazen and Hu, Yihan and Yang, Jiezhi and Higgins, Eric and Brigden, Ryan and Tomizuka, Masayoshi and Zhan, Wei},
  booktitle={CVPR},
  year={2026}
}

@inproceedings{xu2023rayspace,
  title={{SE(3)} Equivariant Convolution and Transformer in Ray Space},
  author={Xu, Yinshuang and Lei, Jiahui and Daniilidis, Kostas},
  booktitle={NeurIPS},
  year={2023}
}

@inproceedings{xie2026urope,
  title={{URoPE}: Universal Relative Position Embedding across Geometric Spaces},
  author={Xie, Yichen and Meng, Depu and Peng, Chensheng and Hu, Yihan and Herau, Quentin and Tomizuka, Masayoshi and Zhan, Wei},
  booktitle={ECCV},
  year={2026}
}

@inproceedings{jang2026raysaspixels,
  title={Rays as Pixels: Learning A Joint Distribution of Videos and Camera Trajectories},
  author={Jang, Wonbong and Liu, Shikun and Sanyal, Soubhik and Perez, Juan Camilo and Ng, Kam Woh and Agrawal, Sanskar and Perez-Rua, Juan-Manuel and Douratsos, Yiannis and Xiang, Tao},
  booktitle={ICML},
  year={2026}
}

@article{kenney2026dppe,
  title={{DPPE}: Rethinking Camera-Based Positional Encoding for Scaling Multi-View Transformers},
  author={Kenney, Shun and Suzuki, Teppei},
  journal={arXiv:2606.31585},
  year={2026}
}

@article{safin2023repast,
  title={{RePAST}: Relative Pose Attention Scene Representation Transformer},
  author={Safin, Aleksandr and Duckworth, Daniel and Sajjadi, Mehdi S. M.},
  journal={arXiv:2304.00947},
  year={2023}
}

@inproceedings{du2023widebaseline,
  title={Learning to Render Novel Views from Wide-Baseline Stereo Pairs},
  author={Du, Yilun and Smith, Cameron and Tewari, Ayush and Sitzmann, Vincent},
  booktitle={CVPR},
  year={2023}
}

@inproceedings{he2020epipolar,
  title={Epipolar Transformers},
  author={He, Yihui and Yan, Rui and Fragkiadaki, Katerina and Yu, Shoou-I},
  booktitle={CVPR},
  year={2020}
}

@inproceedings{wang2022r2l,
  title={{R2L}: Distilling Neural Radiance Field to Neural Light Field for Efficient Novel View Synthesis},
  author={Wang, Huan and Ren, Jian and Huang, Zeng and Olszewski, Kyle and Chai, Menglei and Fu, Yun and Tulyakov, Sergey},
  booktitle={ECCV},
  year={2022}
}

@inproceedings{mildenhall2020nerf,
  title={{NeRF}: Representing Scenes as Neural Radiance Fields for View Synthesis},
  author={Mildenhall, Ben and Srinivasan, Pratul P. and Tancik, Matthew and Barron, Jonathan T. and Ramamoorthi, Ravi and Ng, Ren},
  booktitle={ECCV},
  year={2020}
}

@article{li2026cameraanything,
  title={{CameraAnything}: Refilming Videos with Arbitrary Camera Control},
  author={Li, Yixuan and Zeng, Yanhong and Cheng, Ka Leong and Zhu, Jiayi and Wang, Hanlin and Wang, Wen and Meng, Yihao and Ouyang, Hao and Wang, Qiuyu and Yu, Yue and Wang, ZiDong and Zhang, Yiyuan and Shen, Yujun and Lin, Dahua},
  journal={arXiv:2607.24591},
  year={2026}
}

@inproceedings{bai2025syncammaster,
  title={SynCamMaster: Synchronizing Multi-Camera Video Generation from Diverse Viewpoints},
  author={Bai, Jianhong and Xia, Menghan and Wang, Xintao and Yuan, Ziyang and Fu, Xiao and Liu, Zuozhu and Hu, Haoji and Wan, Pengfei and Zhang, Di},
  booktitle={ICLR},
  year={2025}
}

@inproceedings{kuang2024cvd,
  title={Collaborative Video Diffusion: Consistent Multi-video Generation with Camera Control},
  author={Kuang, Zhengfei and Cai, Shengqu and He, Hao and Xu, Yinghao and Li, Hongsheng and Guibas, Leonidas and Wetzstein, Gordon},
  booktitle={NeurIPS},
  year={2024}
}

@inproceedings{xu2025cavia,
  title={Cavia: Camera-controllable Multi-view Video Diffusion with View-Integrated Attention},
  author={Xu, Dejia and Jiang, Yifan and Huang, Chen and Song, Liangchen and Gernoth, Thorsten and Cao, Liangliang and Wang, Zhangyang and Tang, Hao},
  booktitle={ICML},
  year={2025}
}

@inproceedings{vanhoorick2024gcd,
  title={Generative Camera Dolly: Extreme Monocular Dynamic Novel View Synthesis},
  author={Van Hoorick, Basile and Wu, Rundi and Ozguroglu, Ege and Sargent, Kyle and Liu, Ruoshi and Tokmakov, Pavel and Dave, Achal and Zheng, Changxi and Vondrick, Carl},
  booktitle={ECCV},
  year={2024}
}

@inproceedings{watson20254dim,
  title={Controlling Space and Time with Diffusion Models},
  author={Watson, Daniel and Saxena, Saurabh and Li, Lala and Tagliasacchi, Andrea and Fleet, David J.},
  booktitle={ICLR},
  year={2025}
}

@inproceedings{he2025cameractrlii,
  title={CameraCtrl II: Dynamic Scene Exploration via Camera-controlled Video Diffusion Models},
  author={He, Hao and Yang, Ceyuan and Lin, Shanchuan and Xu, Yinghao and Wei, Meng and Gui, Liangke and Zhao, Qi and Wetzstein, Gordon and Jiang, Lu and Li, Hongsheng},
  booktitle={ICCV},
  year={2025}
}

@inproceedings{yang2024directavideo,
  title={Direct-a-Video: Customized Video Generation with User-Directed Camera Movement and Object Motion},
  author={Yang, Shiyuan and Hou, Liang and Huang, Haibin and Ma, Chongyang and Wan, Pengfei and Zhang, Di and Chen, Xiaodong and Liao, Jing},
  booktitle={SIGGRAPH},
  year={2024}
}

@inproceedings{xiao2025trajectoryattention,
  title={Trajectory Attention for Fine-grained Video Motion Control},
  author={Xiao, Zeqi and Ouyang, Wenqi and Zhou, Yifan and Yang, Shuai and Yang, Lei and Si, Jianlou and Pan, Xingang},
  booktitle={ICLR},
  year={2025}
}

@inproceedings{sun2025dimensionx,
  title={DimensionX: Create Any 3D and 4D Scenes from a Single Image with Controllable Video Diffusion},
  author={Sun, Wenqiang and Chen, Shuo and Liu, Fangfu and Chen, Zilong and Duan, Yueqi and Zhang, Jun and Wang, Yikai},
  booktitle={ICCV},
  year={2025}
}

@inproceedings{huang2026spacetimepilot,
  title={SpaceTimePilot: Generative Rendering of Dynamic Scenes Across Space and Time},
  author={Huang, Zhening and Jeong, Hyeonho and Chen, Xuelin and Gryaditskaya, Yulia and Wang, Tuanfeng Y. and Lasenby, Joan and Huang, Chun-Hao},
  booktitle={CVPR},
  year={2026}
}

@inproceedings{ren2025gen3c,
  title={GEN3C: 3D-Informed World-Consistent Video Generation with Precise Camera Control},
  author={Ren, Xuanchi and Shen, Tianchang and Huang, Jiahui and Ling, Huan and Lu, Yifan and Nimier-David, Merlin and M{\"u}ller, Thomas and Keller, Alexander and Fidler, Sanja and Gao, Jun},
  booktitle={CVPR},
  year={2025}
}

@inproceedings{you2025nvssolver,
  title={NVS-Solver: Video Diffusion Model as Zero-Shot Novel View Synthesizer},
  author={You, Meng and Zhu, Zhiyu and Liu, Hui and Hou, Junhui},
  booktitle={ICLR},
  year={2025}
}

@inproceedings{hou2025camtrol,
  title={Training-free Camera Control for Video Generation},
  author={Hou, Chen and Chen, Zhibo},
  booktitle={ICLR},
  year={2025}
}

@inproceedings{li2025realcami2v,
  title={RealCam-I2V: Real-World Image-to-Video Generation with Interactive Complex Camera Control},
  author={Li, Teng and Zheng, Guangcong and Jiang, Rui and Zhan, Shuigen and Wu, Tao and Lu, Yehao and Lin, Yining and Deng, Chuanyun and Xiong, Yepan and Chen, Min and Cheng, Lin and Li, Xi},
  booktitle={ICCV},
  year={2025}
}

@inproceedings{song2026worldforge,
  title={{WorldForge}: Taming Video Models for 3D and 4D Generation via Zero-Shot Camera Control},
  author={Song, Chenxi and Yang, Yanming and Zhao, Tong and Li, Ruibo and Zhang, Chi},
  booktitle={CVPR},
  year={2026}
}

@inproceedings{lee2026pointtracks,
  title={Generative Video Motion Editing with 3D Point Tracks},
  author={Lee, Yao-Chih and Zhang, Zhoutong and Huang, Jiahui and Wang, Jui-Hsien and Lee, Joon-Young and Huang, Jia-Bin and Shechtman, Eli and Li, Zhengqi},
  booktitle={CVPR},
  year={2026}
}

@inproceedings{zhou2025latentreframe,
  title={Latent-Reframe: Enabling Camera Control for Video Diffusion Model without Training},
  author={Zhou, Zhenghong and An, Jie and Luo, Jiebo},
  booktitle={ICCV},
  year={2025}
}

@article{sun2025worldplay,
  title={WorldPlay: Towards Long-Term Geometric Consistency for Real-Time Interactive World Modeling},
  author={Sun, Wenqiang and Zhang, Haiyu and Wang, Haoyuan and Wu, Junta and Wang, Zehan and Wang, Zhenwei and Wang, Yunhong and Zhang, Jun and Wang, Tengfei and Guo, Chunchao},
  journal={arXiv:2512.14614},
  year={2025}
}

@inproceedings{huang2024vbench,
  title={Vbench: Comprehensive benchmark suite for video generative models},
  author={Huang, Ziqi and He, Yinan and Yu, Jiashuo and Zhang, Fan and Si, Chenyang and Jiang, Yuming and Zhang, Yuanhan and Wu, Tianxing and Jin, Qingyang and Chanpaisit, Nattapol and others},
  booktitle={2024 IEEE/CVF Conference on Computer Vision and Pattern Recognition (CVPR)},
  pages={21807--21818},
  year={2024},
  organization={IEEE}
}

@article{lin2024open,
  title={Open-sora plan: Open-source large video generation model},
  author={Lin, Bin and Ge, Yunyang and Cheng, Xinhua and Li, Zongjian and Zhu, Bin and Wang, Shaodong and He, Xianyi and Ye, Yang and Yuan, Shenghai and Chen, Liuhan and others},
  journal={arXiv preprint arXiv:2412.00131},
  year={2024}
}
